\documentclass[journal]{IEEEtran}

\usepackage[amssymb]{SIunits}
\usepackage{graphicx}
\usepackage{amssymb}
\usepackage{amsthm}
\usepackage{multirow}
\usepackage{enumerate}
\usepackage{enumitem}
\usepackage{wrapfig}
\usepackage{seqsplit}

\usepackage{amsmath}
\usepackage[cmintegrals]{newtxmath}
\usepackage{bm} 
\usepackage{footnote}
\usepackage{threeparttable}
\usepackage{epsfig}
\usepackage{graphicx}
\usepackage{amsmath}
\usepackage{amssymb}
\usepackage{balance}

\usepackage{url}            
\usepackage{booktabs}       
\usepackage{amsfonts}       
\usepackage{nicefrac}       
\usepackage{microtype}      
\usepackage{lipsum}
\usepackage{paralist}
\usepackage{hyperref}
\usepackage{color}

\usepackage{array}
\newcommand{\PreserveBackslash}[1]{\let\temp=\\#1\let\\=\temp}
\newcolumntype{C}[1]{>{\PreserveBackslash\centering}p{#1}}
\newcolumntype{R}[1]{>{\PreserveBackslash\raggedleft}p{#1}}
\newcolumntype{L}[1]{>{\PreserveBackslash\raggedright}p{#1}}

\newcommand{\rom}[1]{\uppercase\expandafter{\romannumeral #1\relax}}

\DeclareMathOperator{\softmax}{softmax}
\DeclareMathOperator{\unitnorm}{unit}

\newcommand{\fref}[1]{Fig.~\ref{#1}}
\newcommand{\sref}[1]{Section~\ref{#1}}
\newcommand{\tref}[1]{Table~\ref{#1}}
\newcommand{\fix}[1]{#1}

\newcommand{\etal}{\textit{et al.}~}
\newcommand{\ie}{{i.e.},~}
\newcommand{\eg}{{e.g.},~}
\newcommand{\etc}{{etc.}~}

\newtheorem*{definition}{Definition}

\newtheorem*{formulation}{Formulation}

\graphicspath{%
    {./figures/}%
}

\ifCLASSOPTIONcompsoc
  \usepackage[nocompress]{cite}
\else
  \usepackage{cite}
\fi
\ifCLASSINFOpdf
\else
\fi
\usepackage{amsmath}
\ifCLASSOPTIONcompsoc
  \usepackage[caption=false,font=footnotesize,labelfont=sf,textfont=sf]{subfig}
\else
  \usepackage[caption=false,font=footnotesize]{subfig}
\fi
\usepackage{url}

\begin{document}

%
\title{\fix{Unsupervised Online Learning for Robotic Interestingness with Visual Memory}}
%
%
%

\ifCLASSOPTIONcompsoc
  \author{Chen~Wang,~
        Yuheng~Qiu,~
        Wenshan~Wang,~
        Yafei~Hu,~
        Seungchan~Kim,~
        and~Sebastian~Scherer
\IEEEcompsocitemizethanks{\IEEEcompsocthanksitem The authors are with the Robotics Institute, Carnegie Mellon University, Pittsburgh, PA 15213, USA. E-mail: {\tt chenwang@dr.com, \{yuhengq, wenshanw, yafeih, seungch2, basti\}@andrew.cmu.edu} \protect \hfil
\IEEEcompsocthanksitem \url{https://github.com/wang-chen/interestingness} \protect \hfil
}
\thanks{Manuscript received September xx, 20xx; revised November xx, 20xx.}
}
\else
\author{Chen~Wang,~
        Yuheng~Qiu,~
        Wenshan~Wang,~
        Yafei~Hu,~
        Seungchan~Kim,~
        and~Sebastian~Scherer
\thanks{
This work was partially sponsored by the ONR grant \#N0014-19-1-2266 and ARL DCIST CRA award W911NF-17-2-0181. The human subject survey was approved under \#2019\_00000522. This paper was presented in part at the European Conference on Computer Vision (ECCV), 2020 \cite{wang2020visual}.
}
\thanks{The authors are with the Robotics Institute, Carnegie Mellon University, Pittsburgh, PA 15213, USA. (e-mail: {\tt chenwang@dr.com, \{yuhengq, wenshanw, yafeih, seungch2, basti\}@andrew.cmu.edu)}\protect \hfil}
\thanks{Source code is available at \url{https://github.com/wang-chen/interestingness}.}
}
\fi

%
%

\markboth{IEEE TRANSACTIONS ON ROBOTICS
}%
{Chen Wang\MakeLowercase{\textit{et al.}}: Unsupervised Online Learning for Robotic Interestingness with Visual Memory}
\IEEEtitleabstractindextext{%
\begin{abstract}
Autonomous robots frequently need to detect ``interesting'' scenes to decide on further exploration, or to decide which data to share for cooperation. These scenarios often require fast deployment with little or no training data. Prior work considers ``interestingness'' based on data from the same distribution. Instead, we propose to develop a method that automatically adapts online to the environment to report interesting scenes quickly. To address this problem, we develop a novel translation-invariant visual memory and design a three-stage architecture for long-term, short-term, and online learning, which enables the system to learn human-like experience, environmental knowledge, and online adaption, respectively. With this system, we achieve an average of 20\% higher accuracy than the state-of-the-art unsupervised methods in a subterranean tunnel environment. We show comparable performance to supervised methods for robot exploration scenarios showing the efficacy of our approach. We expect that the presented method will play an important role in the robotic interestingness recognition exploration tasks.
\end{abstract}

}

\maketitle

\IEEEdisplaynontitleabstractindextext

%
\IEEEpeerreviewmaketitle

\ifCLASSOPTIONcompsoc
  \IEEEraisesectionheading{\section{Introduction}}
\else
  \section{Introduction}
\fi
\label{sec:introduction}

%


%
%
%
%

\begin{figure}[!t]
    \centering
    \includegraphics[width=1.0\linewidth]{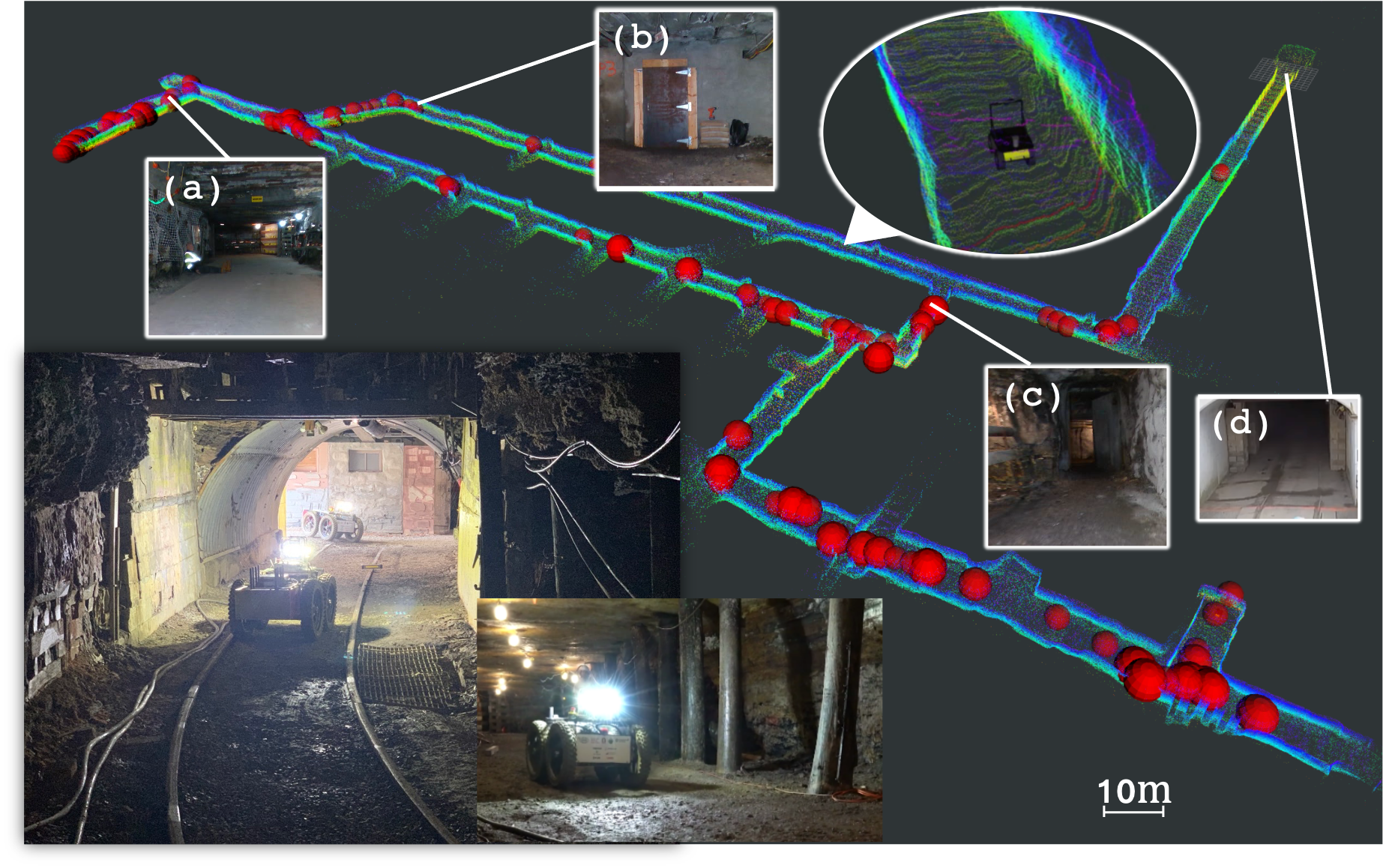}
    \caption{The live interestingness map and our unmanned ground vehicles (UGV) in an exploration task. The red markers indicate the location of detected interesting scenes and the size of the markers reflects the interestingness level. Different from anomaly detection, interestingness recognition needs to find unique and unknown scenes, which means that it has to lose interest in repetitive scenes, thus unsupervised online learning is expected.}
    \label{fig:interestingness-map}
\end{figure}

\begin{figure*}[!t]
    \centering
    \includegraphics[width=1.0\textwidth]{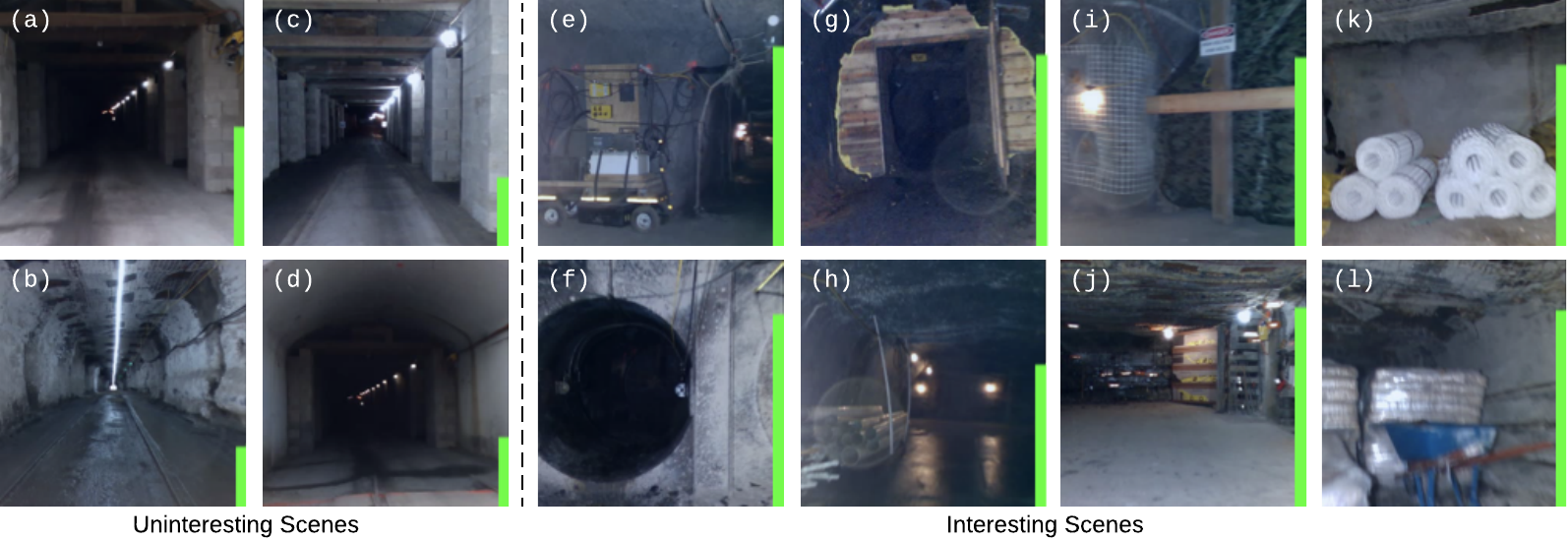}
    \caption{
        Several examples of both uninteresting and interesting scenes in the SubT dataset \cite{subtdata} taken by autonomous robots. The height of green strip located at the right of each image indicates the interestingness level predicted by our unsupervised online learning algorithm when it sees the scene for the first time.}
    \label{fig:motivation}
\end{figure*}

\IEEEPARstart{I}{nteresting} scene recognition is crucial for autonomous exploration, which is one of the most fundamental capabilities of mobile robots \cite{osswald2016speeding}.
It has a significant impact on decision-making and robot cooperation but has received limited attention.
Take the exploration task in \fref{fig:interestingness-map} and \ref{fig:motivation} as an example, finding a narrow passage in \fref{fig:interestingness-map} (c) and a door in \fref{fig:interestingness-map} (b) and \ref{fig:motivation} (g) may affect path planning, while the hole in the wall in \fref{fig:motivation} (f) may attract more attention for future exploration.
However, prior algorithms often have difficulty when they are deployed to unknown environments, as the robots not only have to find interesting scenes but also need to lose interest in repetitive scenes, \ie interesting scenes may become uninteresting during robot exploration after repeatedly observing similar scenes or objects.
For example, a robot may find many things interesting when entering a mine tunnel in \fref{fig:interestingness-map}, while it should quickly lose interest in the repetitive scenes such as tunnel walls, railways, and headlights, although they could and should receive attention for their first appearance.
Nevertheless, recent approaches of interestingness detection \cite{gygli2016analyzing,jiang2013understanding}, saliency detection \cite{zhang2017learning}, anomaly detection \cite{zhao2011online,luo2017revisit}, novelty detection \cite{abati2019latent}, and meaningfulness detection \cite{hasan2016learning} algorithms cannot achieve this type of online adaptation.

To solve this, we propose to establish an \textit{online learning} scheme to search for interesting sites for robotic exploration.
Existing algorithms are heavily dependent on the back-propagation algorithm \cite{rumelhart1988learning} for learning, which is computationally expensive.
To overcome this difficulty, we introduce a novel \textit{translation-invariant 4-D visual memory} to identify and recollect visually interesting scenes.
Since it has no learnable parameters, our algorithm can run very fast.
Human beings have a great capability to guide visual attention and judge the interestingness \cite{constantin2019computational}.
For mobile robots, we found the following two properties necessary to establish a sense of visual interestingness.

\paragraph{Unsupervised}
As shown in \fref{fig:motivation}, interesting scenes in robotic operating environments are often unique and unknown. Therefore, the labels are normally difficult to obtain, but prior research mainly focused on supervised methods \cite{amengual2015review,constantin2019computational} and suffers from prior unseen environments.
In this paper, we hypothesize that a sense of interestingness can be established in an unsupervised manner for mobile robots.

\paragraph{Task-dependent}
In practical applications, we often only know uninteresting scenes before a mission is started. Following the example of the tunnel exploration task in \fref{fig:motivation}, deployment will be more efficient and easier if the robots can be taught what is uninteresting within several minutes. Therefore, we argue that the visual interestingness recognition system should be able to learn from negative samples quickly so an \textit{incremental learning} method is necessary. Note that we expect the model to be capable of learning from negative samples, but that is not necessary for all tasks.

To achieve the properties above, we introduce a three-stage learning architecture for robotic interestingness recognition:

\paragraph{Long-term learning}
In this stage, we expect a model to be trained offline with big data in an unsupervised manner.
\fix{Inspired by that the animals (human beings) acquire new information about the world through mechanisms of learning, and retain the information through mechanisms of memory \cite{goelet1986long}, we expect the model to acquire long-term knowledge from their experiences (big data) and retain the knowledge by freezing learnable weights.} We expect the training time on a single machine to be on the order of days.

\paragraph{Short-term learning}
For task-dependent knowledge, the model should be able to learn from hundreds of uninteresting images in minutes. This can be done before a mission is started and is beneficial to quick robot deployment.

\paragraph{Online learning}
During mission execution, the system should express the top interests in real-time.
\fix{Similar to the learning process of human beings, which is a special culture that provides continuous development of consciousness in ontogenesis \cite{voitovska2019lifelong}, the interestingness should also be updated online, \eg the interest in repetitive scenes should be lost, even if they are not seen in the short-term learning.}

Another important aspect of online learning is that no data leakage is allowed, \ie each frame is processed without using information from its subsequent frames. This is in contrast to prior works \cite{jiang2013understanding,gygli2016analyzing} and datasets \cite{demarty2017predicting}, where interesting frames can only be selected after the entire sequence is processed \cite{grabner2013visual}. Since robots need to respond in real-time, we expect that our algorithms should be capable of adapting quickly. To better measure the capability of online response, we will also introduce a new evaluation metric.

In summary, our contributions are:
\begin{compactitem}
    \item We introduce a simple yet effective three-stage learning architecture for robotic interesting scene recognition, which is crucial for practical applications. We leverage long-term learning to acquire human-like experience, short-term learning for quick robot deployment and acquiring task-related knowledge, and online learning for environmental adaptation and real-time response.
    \item To accelerate the short-term and online learning, we offer a novel 4-D visual memory to replace the back-propagation algorithm. We introduce cross-correlation similarity for translational invariance, which is crucial for perceiving video streams. We also initiate a tangent operator for safe memory writing, which is crucial for incremental learning from negative samples for fast deployment.
    \item To measure the online performance, we introduce a strict evaluation metric, \ie the area under the curve of online precision \eqref{eq:auc-op} to jointly consider precision, recall rate, and online performance.
    \item It is demonstrated that our approach achieves 20\% higher performance than the state-of-the-art algorithms and can find meaningful scenes in practical applications.
\end{compactitem}

A preliminary version of this work was selected for oral presentation \cite{wang2020visual}. In this journal version, we have substantial improvements, including
(1) We provide a mathematical formulation and an overview for our framework in \sref{sec:formulation} to better describe the problem of robotic interestingness recognition and address the difference between anomaly and saliency detection.
(2) We introduce a new balance mechanism, a usage vector, in \sref{sec:writing} for the visual memory to use the space more efficiently, which allows us to achieve comparable accuracy but a faster speed.
We show in \sref{sec:efficiency} that it achieves 69 FPS compared to 14 FPS in the previous version. This makes our algorithm applicable to ultra-low power systems, which is vital for mobile robots.
(3) We add ablation study in \sref{sec:usage-vector} to provide an intuitive explanation for the novel memory usage vector.
(4) We update the overall performance in \sref{sec:comparison-unsupervised} for the SubT dataset  and add comparison with anomaly detection methods including MemAE \cite{gong2019memorizing} and FFP \cite{liu2018future}.
(5) We show that our unsupervised online learning algorithm is even comparable with the supervised method \cite{arandjelovic2016netvlad} in \sref{sec:comparison-supervised}.
(6) We make the new version of source codes publicly available based on the robot operating system (ROS), provide visualization tools for interestingness map in \fref{fig:interestingness-map} for real-time human inspection, and provide the density map to indicate the location of interesting objects in \sref{sec:density-map}.

\begin{figure*}[t]
    \centering
    \includegraphics[width=1.0\textwidth]{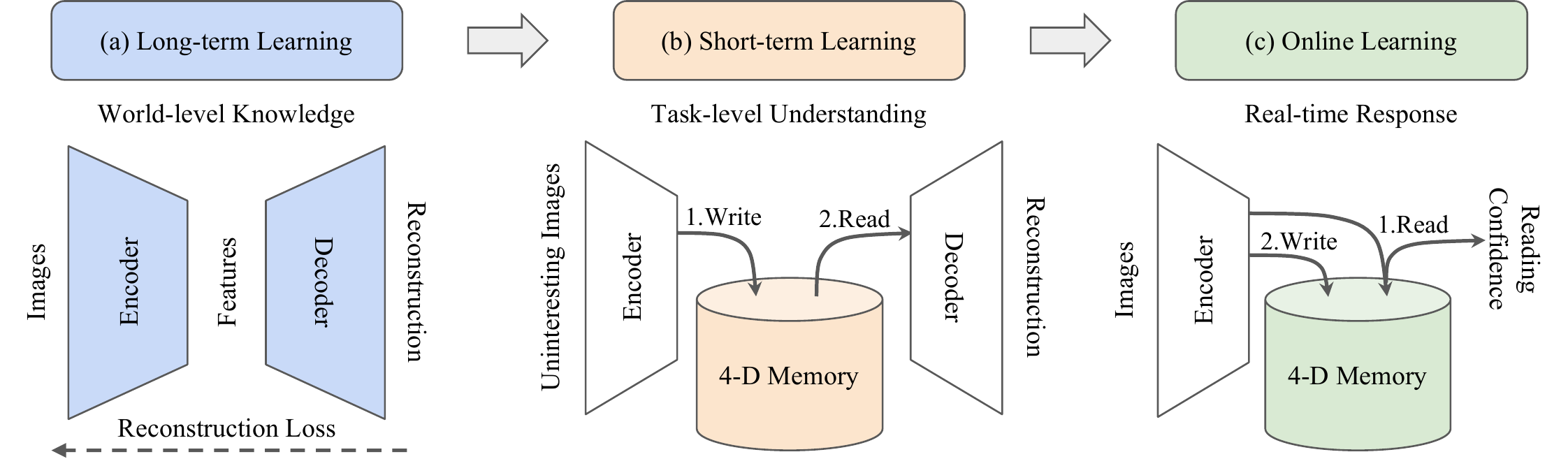}
    \caption{\fix{The three learning stages. (a) During long-term learning, the parameters in both the encoder and decoder are trainable. (b) During short-term learning, the parameters in the encoder and decoder are frozen; the memory writing is performed before reading. (c) During online learning, the parameters in the encoder are frozen; the memory reading is performed before writing.}}
    \label{fig:learning}
\end{figure*}

\section{Related Work}\label{sec:related-work}

A learning system that encodes the three-stage learning architecture for interesting scene recognition has not been achieved yet. Consequently, the problem formulation and performance evaluation will be very different from prior approaches.
\fix{Some existing work on interestingness recognition has different objectives \cite{amengual2015review}, \eg Shen \etal aimed to predict human interestingness on social media \cite{shen2017deep}, which is not suitable for comparison. Therefore, we will also review the related techniques in saliency, anomaly, and novelty detection, since some techniques are also useful for this work.}

The definition of interestingness is subjective, thus human annotations are often averaged over different participants.
To mimic human judgment, prior works have paid much attention to investigate the relationship of human visual interestingness and image features \cite{amengual2015review}.
They are usually motivated by psychological cues and heavily leverage human annotations for training.
This has spawned a large family of supervised learning methods.
For instance, Dhar \etal designed three hand-crafted rules, including attributes of composition, content, and sky-illumination to approximate both aesthetics and interestingness of images \cite{dhar2011high}.
Grabner \etal integrated four features inspired by cognitive concepts to predict interestingness, including raw pixel values, color histogram, HOG feature, and Gist of the scene \cite{grabner2013visual}.
Gygli \etal introduced a set of features computationally capturing three main aspects of visual interestingness, \ie unusualness, aesthetics, and general preferences \cite{gygli2013interestingness}.
Jiang \etal extended image interestingness to video and evaluated hand-crafted visual features for predicting interestingness on the YouTube and Flickr datasets \cite{jiang2013understanding}.
Fu \etal formulated interestingness as a problem of unified learning to rank, which is able to jointly identify human annotation outliers \cite{fu2014interestingness,fu2015robust}.
Although numerous efforts have been devoted, the performance of hand-crafted and feature-based methods is still not satisfactory in unknown environments.

Deep neural networks played more and more significant roles in recent works on interestingness recognition. For example, Gygli \etal introduced VGG features \cite{Simonyan:2015ws} and used a support vector regression model to predict the interestingness of animated GIFs \cite{gygli2016analyzing}.
Chaabouni \etal constructed a customized CNN model to identify salient and non-salient windows for video interestingness recognition \cite{chaabouni2017deep}.
Inspired by a human annotation procedure of pairwise comparison, Wang \etal combined two deep ranking networks \cite{wang2018video} to obtain better performance. This method ranked first in the 2017 interestingness recognition competition \cite{demarty2017mediaeval}.
Shen \etal combined both CNN and LSTM \cite{hochreiter1997long} for feature learning to predict video interestingness \cite{shen2017deep} for media content.

However, the aforementioned methods are highly dependent on human annotation in training, which is labor expensive and unsuitable for real-time response \cite{constantin2019computational}.
Some efforts for unsupervised learning have been made in \cite{ito2012detecting}, where interesting events of videos are detected using the density ratio estimation algorithm with the HOG features \cite{dalal2005histograms}. Nevertheless, the approach cannot adapt well to changing distributions.

In the long-term stage, we introduce an autoencoder \cite{kramer1991nonlinear} for unsupervised learning, which has been widely used for feature extraction in many applications. For example, Hasan \etal  showed that an autoencoder is able to learn regular dynamics and identify irregularity in long-duration videos \cite{hasan2016learning}. Zhang \etal introduced dropout into the autoencoder for pixel-wise saliency detection in images \cite{zhang2017learning}.
Zhao \etal presented a spatio-temporal autoencoder to extract both spatial and temporal features for anomaly detection \cite{zhao2017spatio}.

In order to learn online, we introduce a novel visual memory module into the convolutional neural networks.
Visual memory has been widely investigated in neuroscience \cite{phillips1974distinction}.
While in computer vision, memory-aided neural networks have received limited attention and are used only for several tasks.
For example, Graves \etal presented differentiable neural Turning machines (NTM) \cite{graves2014neural} to couple external memory with recurrent neural networks (RNN).
Later a differentiable neural computer (DNC) was introduced by adding a mechanism into NTM to ensure that the allocated memory does not overlap and interfere \cite{graves2016hybrid}.
Santoro et al. extended NTM and designed a module to efficiently access to the memory \cite{santoro2016meta}.
Gong et al. introduced a memory module into an auto-encoder to remember normal events for anomaly detection \cite{gong2019memorizing}.
Kim et al. introduced the memory network into GANs to remember previously generated samples to alleviate the forgetting problem \cite{kim2018memorization}.
However,  the memories in the above works are defined as flattened vectors, thus the spatial structural information cannot be retained. In this paper, we present a translation-invariant memory module and introduce online learning for robotic interestingness.

\section{Formulation \& Overview}\label{sec:formulation}

To better describe the problem, we define interestingness and the task of interestingness recognition as follows.

\fix{
\begin{definition}[Interestingness]
    In psychology, interestingness is a power of attracting or holding one's attention \cite{interestingness}. In robotics, interestingness is the power of scenes/objects influencing a robotic mission such as exploration and decision-making.
\end{definition}

Intuitively, interestingness is subjective and is often attenuated on repetitive scenes, especially for robotic applications, where unique scenes have higher probabilities to be interesting.
Hence, we next formulate robotic interestingness recognition as a task where temporal information plays an important role.
}

\begin{formulation}
    In the task of interestingness recognition, we will observe an ordered image sequence, \ie $\mathbf{I}_{1:t}=\left[\mathbf{I}_1, \mathbf{I}_2, \dots \mathbf{I}_t\right]$, where each image $\mathbf{I}_i$ is associated with a label $z_i$.
    It is assumed that every item $(\mathbf{I}_t, z_t)$ satisfies $(\mathbf{I}_t, z_t) \sim P_t(\mathbf{I}_{1:t-1}, z_{1:t-1})$, where $P_t$ is a probability distribution describing the interestingness learning task at time $t$.
    The goal of interestingness recognition is to learn a predictor $f(\mathbf{I}_t)$ to associate the ordered image sequence $\mathbf{I}_{1:t}$ with label $z_{1:t}$ such that $(\mathbf{I}_t, z_t) \sim P_t$.
\end{formulation}

It can be seen that the interestingness label $z_t$ is not only related to its associated image $\mathbf{I}_t$, but also related to all its preceding images, hence the order of image sequence $\mathbf{I}_{1:t}$ will also have an impact on their interestingness.
Therefore, its main difference from anomaly detection as well as other tasks such as saliency detection, is the online adaption, since an anomaly event is always anomalous, but recurring interesting scenes may become uninteresting online.
To better learn the predictor $f$, we need a model to take into account of all historical data $\mathbf{I}_{1:t-1}$ for prediction, hence we will introduce a novel visual memory module to better memorize and recall historical scenes.

In this paper, we focus on unsupervised learning for unknown interests, since supervised models normally perform better for known interests. For example, a pre-trained human detector is often preferred in the mission of search and rescue.

To achieve the expected properties for mobile robots, we also need such an online learning algorithm to run very fast on embedded low-power processors.
Therefore, we introduce a simple yet effective three-stage learning architecture in \fref{fig:learning} that consists of a reconstruction and a memory module. We next provide an overview for the three-stage framework and will further explain their behavior in \sref{sec:learning}.

\fix{
The long-term learning stage in \fref{fig:learning} (a) is a reconstruction model serving for the human-like long-term experience.
This stage generally requires a good generalization ability and is to mimic the long-term learning behavior of the human brain \cite{goelet1986long}, in which learned knowledge is not easy to forget.
As mentioned before, we retain the long-term knowledge by freezing learnable weights in the model.
Basically, the model can be any unsupervised model that can learn from experience.}
In the experiments, we use a simple autoencoder for efficiency.

To identify unique interests, we then incorporate the visual memory module in the short-term and online learning in \fref{fig:learning} (b) and \fref{fig:learning} (c).
One important motivation is that the memory module is able to perform fast learning, which makes it very suitable for memorizing and recollecting visited scenes in real-time.
In this way, we are able to identify the interests by comparing the current scene and the most similar visited scenes. Intuitively, the scenes that cannot be well recollected by the memory have higher probabilities to be interesting scenes.

However, this is challenging for existing memory modules including \cite{graves2014neural,graves2016hybrid}, since a live video stream often contains a large amount of redundant information, thus we have to improve the space utilization.
Another challenge is that the robots often move continuously, which will lead the video to contain many translational movements. This requires the memory recollecting to be invariant to translation.
To achieve these properties, we will first define a novel visual memory in \sref{sec:memory} then explain that how this memory can be incorporated into the three-stage learning architecture to identify robotic interestingness.

\section{Visual Memory}\label{sec:memory}

Due to limited storage, we are not able to store all the historical images $\mathbf{I}_i$, but it might be possible to memorize their approximation, i.e., visual features, which will be denoted as $\mathbf{x}_i$.
To retain the structural information of visual inputs, we define visual memory $\mathbf{M}$ as a 4-D tensor, \ie $\mathbf{M}\in\mathbb{R}^{n\times c\times h\times w}$, where $n$ is the number of memory cubes (capacity) and $c$, $h$, and $w$ are the channel, height, and width of each cube, respectively.
A memory has two operations, \ie writing and reading.
\textit{Writing} is to encode visual inputs into the memory, while \textit{reading} is to recall the memory regarding the visual inputs.

\subsection{Memory Writing}\label{sec:writing}

We first present the memory writing in the previous version \cite{wang2020visual}, then show how we can further improve the space usage based on that.
The goal of visual memory is to balance the old knowledge and new knowledge from visual inputs.
Denote the visual inputs at time $t$ as $\mathbf{x}(t)\in\mathbb{R}^{c\times h\times w}$, we can define the writing protocol for the $i$-th memory cube $\mathbf{M}_i$ at time $t$ as
\begin{equation}\label{eq:writing}
\mathbf{M}_i(t) = (1-\mathbf{w}_i(t))\cdot \mathbf{M}_i(t-1) + \mathbf{w}_i(t)\cdot \mathbf{x}(t),
\end{equation}
where $\mathbf{w}_i$ is the i$_{\text{th}}$ element of a weight vector $\mathbf{w}\in[0,1]^n$,
\begin{equation}\label{eq:writing-vector}
\mathbf{w}(t) = \sigma(\gamma_w\cdot\tan(\frac{\pi}{2}\cdot D(\mathbf{x}(t), \mathbf{M}(t-1)))),
\end{equation}
where $\sigma(\cdot)$ is the softmax function and $D(\mathbf{x}, \mathbf{M})$ is a cosine similarity vector, in which the $i$-th element $D_i(\mathbf{x}, \mathbf{M})$ is
\begin{equation}\label{eq:cosine}
D_i(\mathbf{x}, \mathbf{M}) = \frac{\sum(\mathbf{x}\odot \mathbf{M}_i)}{\|\mathbf{x}\|_\mathbf{F}\cdot\|\mathbf{M}_i\|_\mathbf{F}},
\end{equation}
where $\odot$, $\sum$, and $\|\cdot\|_\mathbf{F}$ are element-wise product, elements summation, and Frobenius norm, respectively.
The writing protocol in \eqref{eq:writing} is a moving average, in which the learning speed can be controlled via the writing rate $\gamma_w~(\gamma_w>0)$, so that the training samples can be learned at an expected speed. The effect of wring rate $\gamma_w$ is presented in \sref{sec:losing-interest}.

It's worth mentioning that, to promote the sparsity of memory, we introduce a tangent operator in \eqref{eq:writing-vector} to map the range of cosine similarity $[-1,1]$ in \eqref{eq:cosine} to $[-\inf,\inf]$. Therefore, memory writing can be focused on fewer, but more relevant cubes via the softmax function.
This leads to easier incremental learning compared to \cite{graves2014neural,graves2016hybrid}, which will be explained in \sref{sec:short-term} and \sref{sec:writing-vector}.
Note that giving a very low temperature $T\rightarrow 0$ for the softmax function is also able to map the similarity to $[-\inf,\inf]$; however, the properties of the cosine similarity near to 0 will also be affected.
As a comparison, we have $\tan(x)\approx x$ for $x \rightarrow 0$, which is able to retain the properties of the softmax function.
We notice that writing sparsity is also mentioned in \cite{zhao2011online,luo2017revisit,gong2019memorizing}. However, they are designed for different objectives using different strategies. For instance, to detect an anomaly, Gong \etal \cite{gong2019memorizing} introduced a simple threshold to promote sparsity for reducing reconstruction accuracy.

To further improve the efficiency of memory usage, we argue that the memory cubes $\mathbf{M}_i$ that are less used should receive a higher writing weight $\mathbf{w}_i$. Therefore, we need a memory usage vector $\mathbf{u}(t)\in\mathbb{R}^n$ to represent how much a memory cube is used at time $t$. Intuitively, it can be defined in \eqref{eq:memory-usage} through time recursively, following the writing protocol \eqref{eq:writing} via moving average, which assumes the information of an input $\mathbf{x}(t)$ is 1.
\begin{equation}\label{eq:memory-usage}
 \mathbf{u}(t) =
    \left\{\begin{matrix}
    (1-\mathbf{w}(t))\odot\mathbf{u}(t-1) + \mathbf{w}(t) & t >0\\
    \mathbf{0}_{n\times 1} & t=0
    \end{matrix}\right..
\end{equation}
Hence, we can update the memory writing vector \eqref{eq:writing-vector} to \eqref{eq:writing-update} to address less-used memory cubes.
\begin{equation}\label{eq:writing-update}
  \mathbf{w}_i(t) \leftarrow
  \left\{\begin{matrix}
\unitnorm(\mathbf{w}_i(t) \odot (1-\mathbf{u}_i(t))) & \mathbf{w}_i(t) < \mathbf{u}_i(t)  \\
\mathbf{w}_i(t) & \text{otherwise}
\end{matrix}\right.,
\end{equation}
where $\unitnorm(\cdot)$ is a unit normalization operator.
The writing update function \eqref{eq:writing-update} indicates that a novel visual input, which has a low cosine similarity, tends to use less-used memory spaces.
If all memory spaces are used, \ie $\mathbf{u}(t) = \mathbf{1}_{n\times 1}$, the writing vector in \eqref{eq:writing-vector} will be used.
We notice that a usage vector with a different definition is introduced to DNC \cite{graves2016hybrid} for better processing long sequences such as languages.
In this work, we introduced the usage vector for improving writing efficiency, which will be further demonstrated in \sref{sec:efficiency}.

\subsection{Memory Reading}\label{sec:memory-reading}
Recall that convolutional features (visual inputs) are invariant to small translations due to the concatenation of pooling layers to convolutional layers \cite{goodfellow2016deep}.
To obtain invariance to large translations, people often rely on data augmentation, which is very computationally heavy.
To solve this problem, we introduce an invariance to translation into the memory reading, leveraging that the structural information of visual inputs is retained in memory writing.
Denote 2-D circular translation of the $i$-th memory cube along the horizontal and vertical directions with $(x,y)$ elements at time $t$ as $\mathbf{M}_i^{(x,y)}(t)$. The memory reading $\mathbf{f}(t)\in\mathbb{R}^{c\times h \times w}$ is defined as
\begin{equation}\label{eq:reading}
\mathbf{f}(t) = \sum_{i=1}^{n}\mathbf{r}_i\cdot \mathbf{M}_i^{(x,y)}(t),
\end{equation}
where $\mathbf{r}_i$ is the $i$-th element of reading weight vector $\mathbf{r}\in [0,1]^n$,
\begin{equation}\label{eq:reading-vector}
\mathbf{r} = \sigma(\gamma_r\cdot\tan(\frac{\pi}{2}\cdot S(\mathbf{x}(t), \mathbf{M}(t)))),
\end{equation}
where $\gamma_r>0$ is the reading rate.  The $i$-th element of $S(\mathbf{x}, \mathbf{M})$ is the maximum cosine similarity of $\mathbf{x}$ with $\mathbf{M}_i^{(a,b)}$, where $a=0:h-1$ and $b=0:w-1$ imply all translations.
Intuitively, to find the maximum cosine similarity, we need to repeatedly compute \eqref{eq:cosine} for translated memory cube $h\times w$ times, resulting in high computational complexity.
To solve this problem, we leverage the fast Fourier transform (FFT) to compute the cross-correlation \cite{wang2019kernel}.
Recall that 2-D cross-correlation is the inner-products between the first signal and circular translations of the second signal \cite{wang2018kernel}, and the numerator of cosine similarity in \eqref{eq:cosine} is also the inner products. We can therefore compute the maximum cosine similarity as
\begin{equation}\label{eq:cross-correlation}
S_i(\mathbf{x}, \mathbf{M}) = \frac{\max\mathcal{F}^{-1}(\sum^{c}\hat{\mathbf{x}}^*\odot \hat{\mathbf{M}}_i)}{\|\mathbf{x}\|_\mathbf{F}\cdot\|\mathbf{M}_i\|_\mathbf{F}},
\end{equation}
where $\hat{\cdot}$ is the 2-D FFT, $\cdot^*$ is the complex conjugate, and $\sum^{c}$ is element-wise summation along channel dimension.
The translation $(x,y)$ in \eqref{eq:reading} for the $i$-th memory cube is corresponding to the location of the maximum response,
\begin{equation}
(x,y) = \arg\max _{(a,b)} (\sum^{C}\hat{\mathbf{x}}^*\odot \hat{\mathbf{M}}_i)[a,b].
\end{equation}
In this way, the computational complexity for each memory cube can be reduced from $\mathcal{O}(ch^2w^2)$ to $\mathcal{O}(chw\log hw)$.
Another advantage of translation-invariance in memory reading is that the memory usage becomes more efficient through sharing the cubes with a translational difference, since scene translation is common in the continuous video streams for many robotic applications, \eg robot exploration and object search. This will be further explained in \sref{sec:translation-invariance}.

\section{Learning}\label{sec:learning}

We next present the three-stage learning architecture for visual interestingness that incorporates the memory module into an autoencoder to achieve the important properties mentioned in \sref{sec:introduction} and \ref{sec:formulation}. An overview of the architecture is shown in \fref{fig:learning}, where each of the stages will be explained in the following sections, respectively.

\subsection{Long-term Learning}\label{sec:long-term}

Inspired by the fact that humans have a massive storage capacity \cite{brady2008visual}, we use a reconstruction model in \fref{fig:learning} (a) for long-term learning due to the following reasons.

\paragraph{Unsupervised Knowledge}
A reconstruction model can be trained in an unsupervised way, hence we can collect a massive number of images from the internet or in real-time during execution to training the model without much effort.
This agrees with the objective of long-term learning that is to memorize as many scenes as possible.
In this stage, we still leverage the back-propagation algorithm for training, so a large amount of the knowledge will be `stored' in the learnable parameters, which will be frozen afterward. In this sense, the learned knowledge can be treated as an unforgettable human-like experience, which will be important in the following stages.

\paragraph{Detailed and Semantic}
To precisely reconstruct images, the feature map in the bottleneck layer needs to contain detailed information. On the other hand, since the size of the feature map is normally much smaller than the input images, it has to contain higher-level semantic information.
This is crucial for visual interestingness, since both texture and object-level information may attract one's interests.
Recollect the visual feature $\mathbf{x}_i$ is an approximated version of image $\mathbf{I}_i$, we are able to approximate $P_t$ using the visual memory $\mathbf{M}(t)$, which will play an important role in the following learning stage.

In this paper, we adopt one of the most simplified models, an autoencoder in \fref{fig:learning} (a), to reconstruct images.
We establish the encoder following the architecture of VGG-16 \cite{Simonyan:2015ws} and concatenate 5 deconvolutional blocks \cite{long2015fully} for the decoder.
We tried more recent architectures including ResNet \cite{he2016deep} and MobileNetV2 \cite{sandler2018mobilenetv2} and found the performance is less sensitive to the back-bone models than the visual memory presented in this paper.
Feature maps of an autoencoder are invariant to small translations due to the architecture of CNNs. We leverage the invariance of visual memory to large translations in short-term and online learning.

\subsection{Short-term Learning}\label{sec:short-term}
As aforementioned, we often only know the uninteresting scenes before a robotic mission is started. For known interesting objects, we prefer to use supervised object detectors.
Therefore, we expect that our unsupervised model can be trained \textit{incrementally} with negative labeled samples within several minutes.
This will be helpful for the robots to learn environment-related knowledge and quick robot deployment.

To this end, we develop a short-term learning architecture in \fref{fig:learning} (b). The memory module $\mathbf{M}$ is inserted into the trained reconstruction model, in which all parameters are frozen.
For each sample, the output of the encoder is first written into the memory, then memory reading is taken as inputs of the decoder.
Intuitively, the images cannot be reconstructed well initially, as feature maps are not fully learned by the memory, and memory reading will be different from the encoding outputs. Therefore, we can inspect the reconstruction error to find whether the memory has learned to encode the training samples or not. In the experiment, we used the mean squared error (MSE) loss for both long-term and short-term learning.
\fix{In experiments, we find that the final performance is insensitive to the number of epochs in short-term learning. Therefore, to speed up the deployment process, we early stop short-term learning when either one of the two criteria is satisfied: (1) Visual cues: The decoder provides visually good reconstruction; (2) The MSE loss of the reconstruction is not reduced for 3 epochs. More details about the sensitivity to the short-term learning is presented in \sref{sec:effect-shortterm} and \fref{fig:short-term-epoch}.}

The memory learning is very fast since it has no learnable parameters. It also has several advantages.
Recall that the gradient descent algorithms cannot be directly applied to neural networks for incremental learning, since all trainable parameters are changed during training, leading the model to be biased towards the augmented data (new negative labeled data), resulting in the forgetting of previously learned knowledge.
This phenomenon is also called ``catastrophic forgetting'' in continual learning \cite{lopez2017gradient}.
Although we can fine-tune the model on the entire data, which takes the learned parameters from long-term learning as an initialization, it is too computationally expensive and cannot meet the requirements for short-term learning.
Nevertheless, memory learning is able to solve this problem inherently.
One reason is that we introduce the tangent operator in \eqref{eq:writing-vector} to promote writing sparsity, hence fewer memory cubes are affected, resulting in safer and faster incremental learning. The effects of sparse writing and short-term learning will be shown in \sref{sec:effect-writing} and \sref{sec:effect-shortterm}, respectively.

\subsection{Online Learning}\label{sec:online}

Online learning is one of the most important capabilities for a real-time visual interestingness recognition system, as human feelings always keep changing according to  environments and experiences.
Moreover, people tend to lose interest when repeatedly observing the same objects or exploring the same scenes, which is very common in a video stream from a mobile robot.
Therefore, we aim to establish such an online learning architecture for real-time robotic systems in \fref{fig:learning} (c), where only the frozen encoder and memory are involved.

\fix{
Different from short-term learning, in online learning, the inputs are video stream and memory reading is performed before writing.
Intuitively, if unknown scenes or objects appear suddenly, the confidence of memory reading will become lower than before, which can be treated as a new interest.}
Since the new scenes or objects are then written into the memory, their reading confidence level will become higher in the subsequent images. Therefore, the model will learn to lose interest in repetitive scenes once the scene is remembered by the memory. In this sense, the visual interestingness distribution $P_t(\mathbf{I}_{1:t}, z_{1:t})$ should be negatively correlated with the memory reading confidence. 
In practice, we adopt averaged (over feature channel) cosine similarity of the memory reading and visual inputs to approximate the reading confidence.

A large translation often happens during robot exploration, hence an invariance to large translations introduced in \eqref{eq:reading} is able to further reduce memory consumption and improve the system robustness. This will be demonstrated in \sref{sec:effect-translation} and \sref{sec:translation-invariance}. Instead of selecting interesting frames after processing an entire sequence \cite{constantin2019computational}, we need a few control variables that can be easily adjusted for different applications.
For example, a hyper-parameter writing rate $\gamma_w$ controlling the rate of losing interest will be useful for objects search. This will be further demonstrated in \sref{sec:losing-interest}.

\section{Experiments}\label{sec:experiments}

We will first introduce a new metric to better evaluate the performance of online learning, then the performance on two difficult datasets will be presented.
The effects of online learning, writing sparsity, translational invariance, and short-term learning will also be presented, respectively.

\subsection{Evaluation Metric}

Prior research typically only focused on the precision or recall rate and is not able to capture the online response of interestingness.
Moreover, data leakage often happens in prior methods \cite{jiang2013understanding,gygli2016analyzing} and datasets \cite{demarty2017predicting}, where interesting frames are selected only after the entire sequence is processed \cite{grabner2013visual}.
To solve the problems, we introduce a new metric, \ie \textit{area under the curve of online precision} (AUC-OP) to evaluate one frame without using the information from its subsequent frames.

This metric is stricter and jointly considers online response, precision, and recall rate.
Intuitively, if $K$ frames of a sequence are labeled as interesting in the ground truth, an algorithm is perfect if its top K interesting frames are the same as the ground truth.
Considering a sequence $\mathbf{I}_{1:N}$, we take an interestingness prediction $f(\mathbf{I}_t)$ as a true positive (interesting) if and only if $f(\mathbf{I}_t)$ ranks in the top $K_{t,n}$ among a subsequence $p(\mathbf{I}_{t-n+1~:~t})$,  where $K_{t,n}$ is the number of interesting frames in the ground truth.
Note that the subsequence $\mathbf{I}_{t-n+1~:~t}$ only contains frames before $\mathbf{I}_t$, as data leakage is not allowed in the online response.
Therefore, we may calculate an online precision score for the subsequences (with length $n$) as
\begin{equation}
    s(n) = \frac{\sum\text{TP}}{\sum\text{TP}+\sum\text{FP}},
\end{equation}
where TP and FP denote the number of true positives and false positives, respectively.
Note that all true positives rank in the top $K_{t,n}$, which means that no false negative is allowed. Recollect that a recall rate can be calculated as
$r=\nicefrac{\sum\text{TP}}{\left(\sum\text{TP}+\sum\text{FN}\right)}$, where FN is the number of false negatives.
This means that the proposed online precision score $s(n)$ requires a $100\%$ recall rate.
For a better comparison, we often accept true positive predictions with ranking in the top $\delta \cdot K_{t,n}$, where $\delta \geq 1$.
Therefore, the overall performance that jointly considers online performance, precision, and recall rate is the area under the curve of the online precision $s(\frac{n}{N}, \delta)$, where $\frac{n}{N}\in(0,1]$. This includes all subsequences since length $n\in[1,N]$, hence it can be written as an integral
\begin{equation}\label{eq:auc-op}
    T(\delta) =\int_{0}^{1}s(x, \delta) \mathrm{d} x,
    \tag{AUC-OP}
\end{equation}
where $x=\frac{n}{N}$. In practice, we often allow some false negatives and $\delta=2$ is recommended for most exploration tasks.
Note that most robotic systems require real-time response, thus their performance on \ref{eq:auc-op} is vital. However, the evaluation of real-time response is often ignored by existing works.

\begin{table}[!t]
    \centering
    \begin{threeparttable}
        \caption{Statistics of the SubT dataset \cite{subtdata}.}
        \label{tab:subt}
        \begin{tabular}{C{0.18\linewidth}C{0.19\linewidth}C{0.19\linewidth}C{0.19\linewidth}}
            \toprule
            Sequence & Length  & Normal & Difficult \\\midrule
            \rom{1} & 53.1 \minute & 11.11\% & 2.76\% \\
            \rom{2} & 55.7 \minute & 15.07\% & 4.49\% \\
            \rom{3} & 79.4 \minute &  9.37\% & 3.02\% \\
            \rom{4} & 80.0 \minute & 17.51\% & 4.29\% \\
            \rom{5} & 59.0 \minute & 24.52\% & 4.07\% \\
            \rom{6} & 57.5 \minute & 22.77\% & 3.30\% \\
            \rom{7} & 83.0 \minute & 11.04\% & 3.21\% \\
            Overall & 467.7 \minute & 15.49\% & 3.58\% \\
            \bottomrule
        \end{tabular}
        \label{tab:subtf}
        \begin{tablenotes}[normal,flushleft]
            \item ``Normal'' and ``Difficult'' indicates that the percentage of frames that are labelled as interesting by at least 1 or 2 subjects, \fix{respectively}.
        \end{tablenotes}
    \end{threeparttable}
\end{table}

\begin{figure*}[t]
    \centering
    \includegraphics[width=1.0\textwidth]{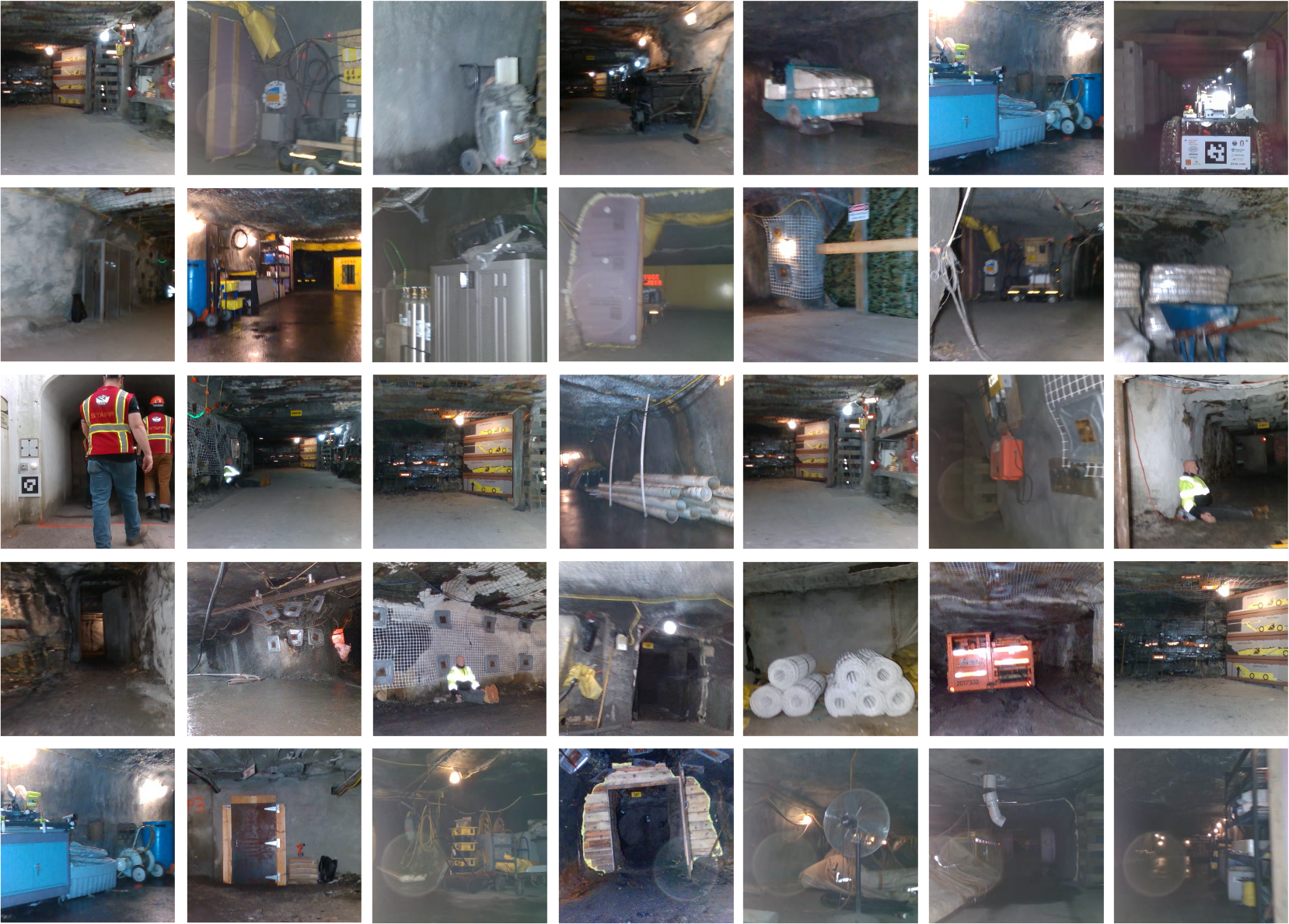}
    \caption{Examples of the detected interesting scenes by our unsupervised online learning algorithm.}
    \label{fig:interestingness}
\end{figure*}

\subsection{Implementation}
Our algorithm is implemented in the robot operating system (ROS) using the PyTorch library \cite{paszke2017automatic}.
The memory is initialized with a uniform distribution described in \cite{he2015delving}. The reading and writing rates are set as $\gamma_r=\gamma_w=5$.
In the previous version \cite{wang2020visual}, a memory capacity of 1000 is adopted.
In this paper, we can use a smaller memory capacity of 100 to achieve a similar performance due to the improved efficiency of space usage.

\subsection{Dataset}
In long-term learning, we perform unsupervised training with the COCO dataset \cite{lin2014microsoft}.
In short-term and online learning, we will test our online robotic interestingness recognition system on challenging environments, including the underground and air scenarios.
They are the SubT dataset \cite{subtdata} for the unmanned ground vehicles (UGV) and the Drone Filming dataset \cite{wang2019improved} for unmanned aerial vehicles (UAV). The SubT dataset will be used for quantitative evaluation, while the Drone Filming dataset will be mainly for qualitative evaluation.

The SubT dataset is based on the Defense Advanced Research Projects Agency (DARPA) Subterranean Challenge (SubT) Tunnel Circuit.
In this challenge, the competitors are expected to build robotic systems to autonomously search and explore the subterranean environments. The environments pose significant challenges, including a lack of lighting, lack of GPS and wireless communication, dripping water, thick smoke, and cluttered or irregularly shaped environments.
Each of the tunnels has a cumulative linear distance of 4-8 \kilo\meter.
The dataset listed in \tref{tab:subt} contains seven videos ($1\hour$) recorded by two fully autonomous UGV during the competition from Team Explorer \cite{teamexplorer}, who won the first place \cite{subtresult} at this event.
\fix{Each sequence is evaluated by at least three volunteers (robotics engineers and researchers), who are expected to select the frames that could be interesting for a robotic mission. The examples include (a) a scene/object appears for the first time; (b) a scene/object has a sudden change; (c) objects and scenes are mismatched; (d) scenes or objects appear in a new viewpoint.}
It can be seen in \tref{tab:subt} that the SubT dataset is very challenging, as human annotation varies a lot, \ie only 15.5\% and 3.6\% of the frames are labeled as interesting by at least one and two subjects, respectively. Since predicting the category with fewer samples is more difficult, we name the two categories in \tref{tab:subt} as normal and difficult, respectively.

The Drone Filming dataset is recorded by quadcopters during autonomous aerial filming \cite{wang2019improved} and publicly available \cite{dronefliming}.
It also contains challenging environments including intensive light changes, severe vibrations, and motion blur, \etc
Different from other scenarios such as surveillance cameras in anomaly detection, robotic visual systems pose extra challenges due to fast background changes, limited computational resources, and dangerous operating environments to which human beings cannot get access.
We next show that our method is suitable for these scenarios and even better than supervised methods.

\begin{table}[t]
    \centering
    \begin{threeparttable}
        \caption{Runtime of different learning stages.}
        \label{tab:efficiency}
        \begin{tabular}{C{0.13\linewidth}C{0.27\linewidth}C{0.27\linewidth}}
            \toprule
            Method       & Short-term Learning & Online Learning \\
            \midrule
            \cite{wang2020visual}$^\dagger$ & 10 \minute & 72.01 \milli\second/frame  \\
            Ours & \textbf{3} \minute & \textbf{14.58} \milli\second/frame \\
            \bottomrule
        \end{tabular}
        \begin{tablenotes}[normal,flushleft]
            \item  $^\dagger$\cite{wang2020visual} is the conference version of this work.
        \end{tablenotes}
    \end{threeparttable}
\end{table}

\subsection{Efficiency}\label{sec:efficiency}

\fix{We conducted the efficiency test on a single Nvidia GPU of GeForce GTX 1080Ti.}
In long-term learning, it takes about 3 days to perform unsupervised training on the COCO dataset \cite{lin2014microsoft}.
The running time for short-term and online learning is presented in the \tref{tab:efficiency}, together with the previous version of this work \cite{wang2020visual}.
For short-term learning, our model only takes about 3 minutes to learn 912 uninteresting images in the SubT dataset, which is feasible for deployment purposes of most practical applications.
For online learning, it runs about 14.58 \milli\second~per frame, which is feasible for a real-time interestingness recognition system (Real-time means as fast as a human brain, \ie 100 \milli\second/frame \cite{potter1969recognition}).
It can be seen in \tref{tab:efficiency} that we achieve a much faster running speed than the previous version. The main reason is that we introduce the usage vector \eqref{eq:memory-usage} to dramatically improve the efficiency of space usage so that we can use the smaller memory capacity of $n=100$ instead of 1000 to achieve similar performance.
This is important for mobile robots since it makes our method applicable to ultra-low power processors. We further provide an intuitive explanation for the effects of usage vector in \sref{sec:usage-vector}.

\begin{table*}[!t]
    \centering
    \begin{threeparttable}
        \caption{Comparison with unsupervised methods on the SubT dataset.$^\dagger$}
        \label{tab:comparison}
        \begin{tabular}{C{0.1\linewidth}|C{0.041\linewidth}C{0.041\linewidth}C{0.041\linewidth}|C{0.041\linewidth}C{0.041\linewidth}C{0.041\linewidth}|C{0.041\linewidth}C{0.041\linewidth}C{0.041\linewidth}|C{0.041\linewidth}C{0.041\linewidth}C{0.041\linewidth}}
            \toprule
            \multicolumn{13}{c}{Normal Category} \\ \midrule
            \multirow{2}{*}{Sequence} & \multicolumn{3}{c|}{w/o invariance} & \multicolumn{3}{c|}{MemAE \cite{gong2019memorizing}} & \multicolumn{3}{c|}{FFP \cite{liu2018future}$^\ddagger$} & \multicolumn{3}{c}{Ours} \\
            & $\delta=1$  & $\delta=2$  & $\delta=3$  & $\delta=1$ & $\delta=2$ & $\delta=3$ & $\delta=1$  & $\delta=2$  & $\delta=3$  & $\delta=1$ & $\delta=2$ & $\delta=3$\\
            \midrule
            \rom{1} & 0.279 & 0.496 & 0.590 & 0.382 & 0.494 & 0.584 & 0.575 & 0.750 & 0.855 & \textbf{0.614} & \textbf{0.783} & \textbf{0.872} \\
            \rom{2} & 0.359 & 0.613 & 0.805 & 0.405 & 0.624 & 0.803 & 0.616 & 0.785 & 0.901 & \textbf{0.632} & \textbf{0.841} & \textbf{0.933} \\
            \rom{3} & 0.186 & 0.239 & 0.293 & 0.235 & 0.311 & 0.377 & 0.610 & 0.805 & 0.885 & \textbf{0.648} & \textbf{0.829} & \textbf{0.898} \\
            \rom{4} & 0.240 & 0.382 & 0.520 & 0.288 & 0.422 & 0.541 & 0.521 & 0.748 & 0.936 & \textbf{0.626} & \textbf{0.845} & \textbf{0.966} \\
            \rom{5} & 0.423 & 0.658 & 0.843 & 0.418 & 0.642 & 0.851 & \textbf{0.767} & 0.921 & 0.968 & {0.746} & \textbf{0.922} & \textbf{0.979} \\
            \rom{6} & 0.405 & 0.652 & 0.832 & 0.515 & 0.754 & 0.863 & \textbf{0.743} & \textbf{0.897} & 0.950 & 0.706 & 0.879 & \textbf{0.957} \\
            \rom{7} & 0.416 & 0.533 & 0.638 & 0.398 & 0.528 & 0.607 & 0.521 & 0.680 & 0.835 & \textbf{0.540} & \textbf{0.759} & \textbf{0.846} \\
            \midrule
            Overall & 0.330 & 0.510 & 0.645 & 0.377 & 0.539 & 0.661 & 0.622 & 0.798 & 0.904 & \textbf{0.645} & \textbf{0.837} & \textbf{0.922} \\ \midrule
            \multicolumn{13}{c}{Difficult Category} \\ \midrule
            \rom{1} & 0.136 & 0.164 & 0.245 & 0.227 & 0.307 & 0.363 & \textbf{0.392} & \textbf{0.565} & \textbf{0.672 }& 0.389 & 0.563 & 0.669 \\
            \rom{2} & 0.285 & 0.340 & 0.411 & 0.281 & 0.404 & 0.453 & 0.422 & 0.537 & 0.647 & \textbf{0.476} & \textbf{0.605} & \textbf{0.684} \\
            \rom{3} & 0.122 & 0.152 & 0.173 & 0.137 & 0.200 & 0.229 & 0.412 & 0.621 & 0.707 & \textbf{0.485} & \textbf{0.659} & \textbf{0.700} \\
            \rom{4} & 0.165 & 0.205 & 0.240 & 0.162 & 0.203 & 0.268 & 0.241 & 0.383 & 0.525 & \textbf{0.336} & \textbf{0.523} & \textbf{0.691} \\
            \rom{5} & 0.176 & 0.259 & 0.343 & 0.115 & 0.206 & 0.253 & 0.330 & 0.568 & 0.628 & \textbf{0.411} & \textbf{0.586} & \textbf{0.715} \\
            \rom{6} & 0.219 & 0.315 & 0.381 & 0.244 & 0.341 & 0.388 & 0.299 & 0.578 & 0.704 & \textbf{0.402} & \textbf{0.571} & \textbf{0.661} \\
            \rom{7} & 0.378 & 0.445 & 0.476 & 0.277 & 0.408 & 0.491 & 0.369 & 0.555 & 0.614 & \textbf{0.447} & \textbf{0.542} & \textbf{0.632} \\
            \midrule
            Overall & 0.212 & 0.268 & 0.324 & 0.206 & 0.296 & 0.349 & 0.352 & 0.544 & 0.643 & \textbf{0.421} & \textbf{0.578} & \textbf{0.679} \\
            \bottomrule
        \end{tabular}
        \begin{tablenotes}[normal,flushleft]
            \item  $^\dagger$We report the performance of \ref{eq:auc-op} using $\delta=$ $1$, $2$, $3$, which is a little stricter than previous version \cite{wang2020visual} using $\delta=$ $1$, $2$, $4$.
            \item $^\ddagger$FFP is fine-tuned on mixed sequences containing Sequence \rom{1}, while ours is an unsupervised online learning method without any fine-tuning.
        \end{tablenotes}
    \end{threeparttable}
\end{table*}

\begin{table}[t]
    \centering
    \begin{threeparttable}
        \fix{
        \caption{Comparison on SubT dataset using traditional metrics.}
        \label{tab:prc}
        \begin{tabular}{C{0.16\linewidth}C{0.085\linewidth}C{0.085\linewidth}C{0.085\linewidth}C{0.085\linewidth}C{0.085\linewidth}C{0.085\linewidth}}
            \toprule
            \multirow{2}{*}{Metric} & \multicolumn{3}{c}{Normal Category} & \multicolumn{3}{c}{Difficult Category} \\
                              & MemAE  & FFP  & Ours  & MemAE  & FFP  & Ours \\
            \midrule
            Precision & 0.344 & 0.639 & \textbf{0.654} & 0.181 & 0.366 & \textbf{0.419} \\
            AUC-ROC$^\dagger$ & 0.477  &  0.757  &  \textbf{0.794} & 0.455  &  0.739  &  \textbf{0.754} \\
            \bottomrule
        \end{tabular}
        \begin{tablenotes}[normal,flushleft]
              \item  $^\dagger$The Area Under the Curve of Receiver Characteristic Operator.
        \end{tablenotes}
        }
    \end{threeparttable}
\end{table}

\subsection{Performance}

To the best of our knowledge, robotic visual interestingness recognition is currently underexplored and existing methods have poor performance in this scenario.
To demonstrate the effectiveness of our method, we will compare both unsupervised methods and weakly supervised methods.

\subsubsection{Comparison with Unsupervised Methods}\label{sec:comparison-unsupervised}

We first compare with the unsupervised methods for anomaly detection, i.e., future frame prediction (FFP) \cite{liu2018future} and MemAE \cite{gong2019memorizing}.
To detect an anomaly, FFP introduces temporal constraint into video prediction, while MemAE introduces a memory module.
The overall performance of \ref{eq:auc-op} of our method is presented in \tref{tab:comparison}.
Compared to FFP, our method achieves an average of 2.3\%,3.9\%, and 1.8\% higher accuracy in the normal category and 6.9\%, 3.4\%, and 3.6\% higher accuracy in the difficult category for $\delta=1$, $2$, $3$, respectively.
Note that to obtain better performance, we adopt a pre-trained model from FFP \cite{liu2018future} and fine-tune it on mixed subterranean videos including Sequence \rom{1} from the SubT dataset.
This might be the reason that FFP performs a little better on Sequence \rom{1} than our method.
\fix{
Some interesting scenes predicted by our method are presented in \fref{fig:interestingness}, in which it can be seen that many interesting and meaningful scenes, including doors and intersections, are identified correctly.}

It can be seen that MemAE \cite{gong2019memorizing} has poor performance on the SubT dataset, which is just a little higher than our memory model without translational invariance (w/o invariance). This is not surprising since MemAE is designed for anomaly detection, but not interestingness recognition. One main difference is that video stream in anomaly detection is mostly from a static surveillance camera, thus the background doesn’t change, while in robotic interestingness recognition, there are many fast translational movements. Moreover, MemAE cannot perform online learning to adapt to challenging repetitive environments, hence many false positives are produced.

\fix{
To better illustrate the evaluation, we next report the performance using other widely-used metrics, including precision and the area under the curve (AUC) of receiver operating characteristic (ROC).
The averaged performance on all sequences is listed in \tref{tab:prc}. It can be seen that our method still achieves the best performance.
Specifically, our method achieves 1.5\% and 5.3\% higher performance in precision and 3.7\% and 1.5\% higher performance in AUC-ROC than FFP in the normal and difficult categories, respectively.
This indicates that our method is still effective even using the traditional metrics.
}

\begin{figure*}[!t]
    \centering
    \subfloat[The Normal Category.]
    {\includegraphics[width=0.4\linewidth]{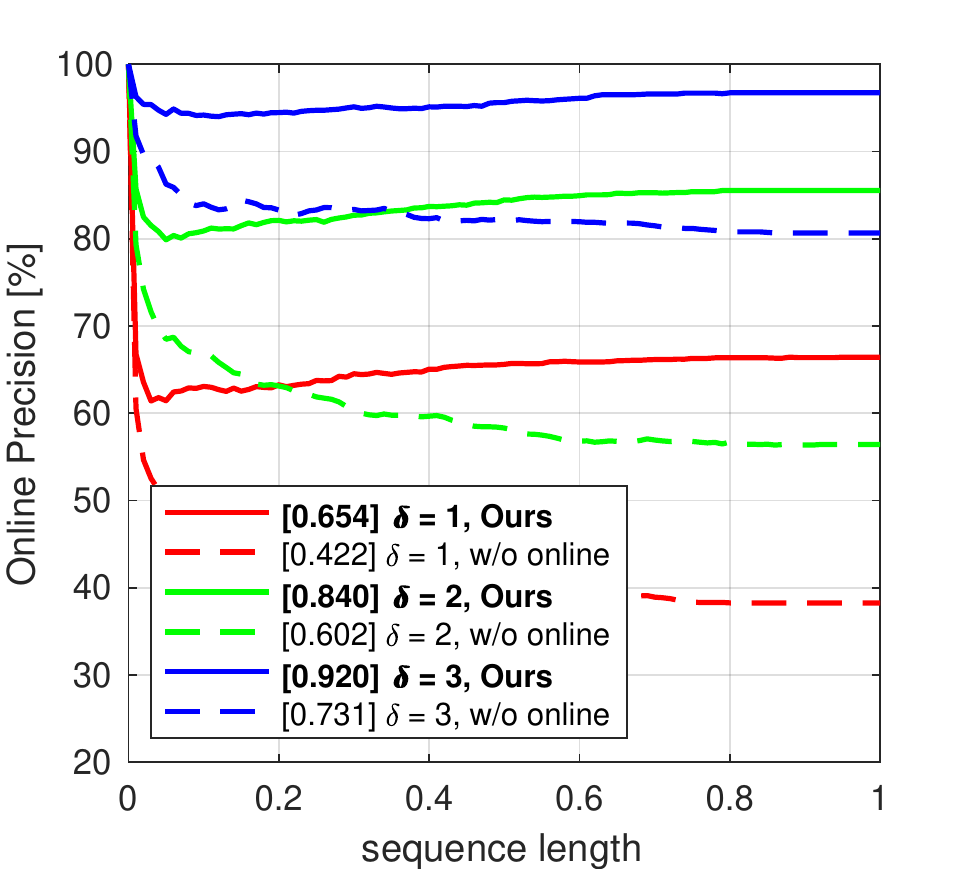}%
        \label{fig:subt-union}
    }
    \hfil
    \subfloat[The Difficult Category.]
    {\includegraphics[width=0.4\linewidth]{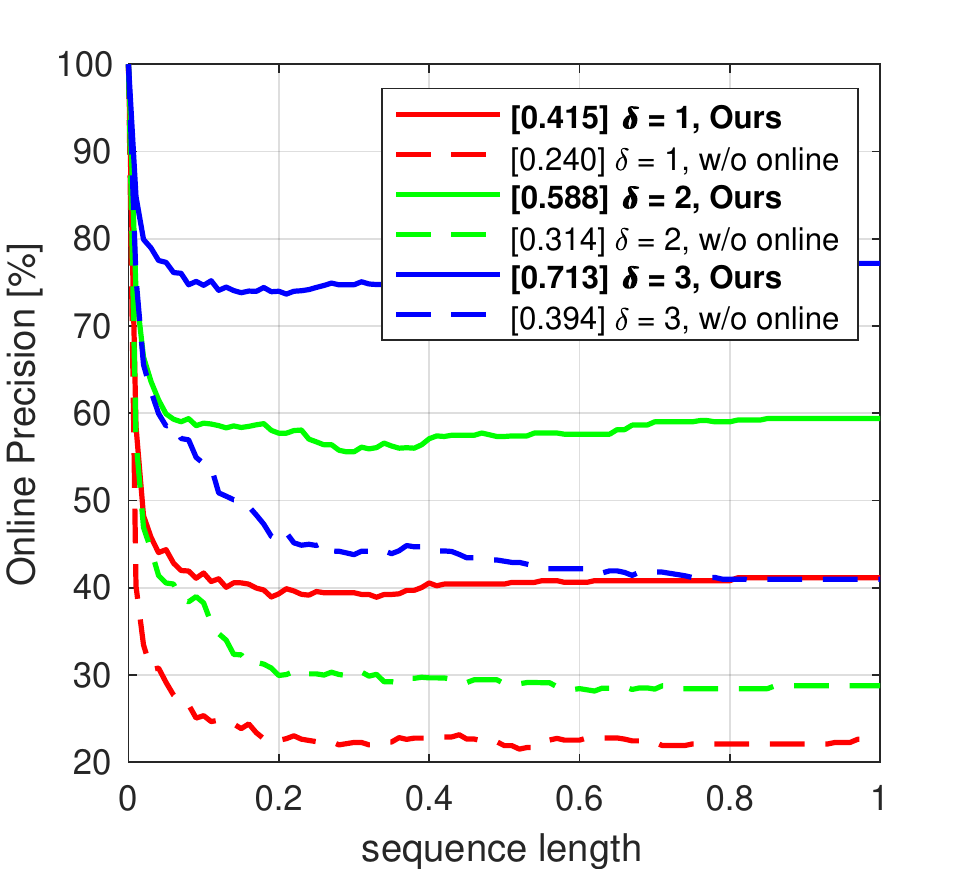}%
        \label{fig:subt-intersection}
    }
    \caption{The overall performance of the \textit{area under the curve of online precision} \ref{eq:auc-op} with and without (w/o) online learning on the SubT dataset. It can be seen that our online learning method achieves an average of 10\%-20\% higher overall performance than the one without online learning in both categories.}
\end{figure*}

\subsubsection{Comparison with Weakly Supervised Methods}\label{sec:comparison-supervised}

We next present the comparison with a weakly supervised method for place recognition, NetVLAD \cite{arandjelovic2016netvlad}.
\fix{
It is inspired by the image representation of vector of locally aggregated descriptor (VLAD) \cite{jegou2010aggregating}, that commonly used in image retrieval.}
The VLAD layer is pluggable into any CNN architecture and has a good generalization ability.
It has been widely used in many place recognition methods such as LPD-Net \cite{liu2019lpd} and PCAN \cite{zhang2019pcan}.
In the experiments, we adopt the architecture of VGG-16 \cite{Simonyan:2015ws} as its feature extractor and train the network with triplet loss \cite{balntas2016learning} using the Adam optimizer \cite{kingma2014adam}, where a learning rate of 0.0001 is adopted. We take the first sequence of the SubT dataset for training, while test on all the other sequences. The training process takes about 2 days to converge.

\begin{table}[t]
    \centering
    \begin{threeparttable}
        \caption{Comparison with supervised methods on the SubT dataset.$^\dagger$}
        \label{tab:unsupervised}
        \begin{tabular}{C{0.13\linewidth}C{0.085\linewidth}C{0.085\linewidth}C{0.085\linewidth}C{0.085\linewidth}C{0.085\linewidth}C{0.085\linewidth}}
            \toprule
            \multirow{2}{*}{Methods} & \multicolumn{3}{c}{Normal Category} & \multicolumn{3}{c}{Difficult Category} \\
            & $\delta=1$  & $\delta=2$  & $\delta=4$  & $\delta=1$ & $\delta=2$ & $\delta=4$\\
            \midrule
            NetVLAD$^\ddagger$ & 0.586 &  \textbf{0.851}  &  0.956 & 0.346  &  \textbf{0.688}  &  \textbf{0.812} \\
            ours & \textbf{0.650}   & 0.846   & \textbf{0.964} & \textbf{0.426}  &  0.581  &  0.749 \\
            \bottomrule
        \end{tabular}
        \begin{tablenotes}[normal,flushleft]
              \item  $^\dagger$The reported performance is an overall average on the Sequence \rom{2} to \rom{7}.
              \item  $^\ddagger$NetVLAD \cite{arandjelovic2016netvlad} is pre-trained on Sequence \rom{1} of the SubT dataset, while ours is still an online unsupervised learning method.
        \end{tablenotes}
    \end{threeparttable}
    \label{tab:unsupervised-comparison}
\end{table}

For fair comparison, we only list the overall performance on the test Sequence \rom{2} to \rom{7}  in \tref{tab:unsupervised} using the metric of \eqref{eq:auc-op}. 
\fix{
It is not surprising that the supervised method NetVLAD outperforms our method in some cases.
However, our method is trained online without labels and still achieves an average of $6.4\%$ and $8.0\%$ higher precision in the most strict case of $\delta=1$, respectively.}
This further verifies the effectiveness and efficiency of our method, especially considering that NetVLAD requires a long time in short-term learning, needs lots of human efforts for annotation, and has to be pre-trained in similar test environments.

\subsection{Effect of Different Modules}
We next present the effects of the proposed modules and learning stages including online learning, writing sparsity, translational invariance, and short-term learning.

\subsubsection{Effect of Online Learning}
Online learning is able to remove many repetitive scenes, thus it can reduce the number of false positives. The curve of online precision for the normal category and difficult category are presented in \fref{fig:subt-union} and \fref{fig:subt-intersection}, where the overall performance of \ref{eq:auc-op} is shown in the associated square brackets.
Our model achieves an average of 23.2\%, 23.8\%, and 18.9\% higher performance than the model without online learning (w/o online) in the normal category for $\delta=$ 1, 2, and 3, respectively.
In the difficult category, our model still has a higher overall performance of 17.5\%, 27.4\%, and 31.9\% for $\delta=$ 1, 2, and 3, respectively, which further confirms the importance of online learning.

\begin{table}[t]
    \centering
    \begin{threeparttable}
        \caption{Effects of the proposed modules on SubT (AUC-OP).}
        \label{tab:overall}
        \begin{tabular}{ccccccc}
            \toprule
            \multirow{2}{*}{Methods} & \multicolumn{3}{c}{Normal Category} & \multicolumn{3}{c}{Difficult Category} \\
            & $\delta=1$  & $\delta=2$  & $\delta=4$  & $\delta=1$ & $\delta=2$ & $\delta=4$\\
            \midrule
            w/o sparsity$^\dagger$ & 0.437 & 0.633  & 0.846 & 0.260 & 0.373 & 0.523 \\
            w/o invariance & 0.330 & 0.510  & 0.752 & 0.212 & 0.268 & 0.379 \\
            w/o short-term & 0.508 & 0.711 & 0.913 & 0.329 & 0.450 & 0.621 \\
            ours & \textbf{0.654} & \textbf{0.840} & \textbf{0.957} & \textbf{0.415} & \textbf{0.588} & \textbf{0.763} \\
            \bottomrule
        \end{tabular}
        \begin{tablenotes}[normal,flushleft]
            \item  $^\dagger$``w/o sparsity''  means without the proposed sparse operator.
        \end{tablenotes}
    \end{threeparttable}
\end{table}

\begin{figure*}[!t]
    \centering
    \includegraphics[width=0.95\linewidth]{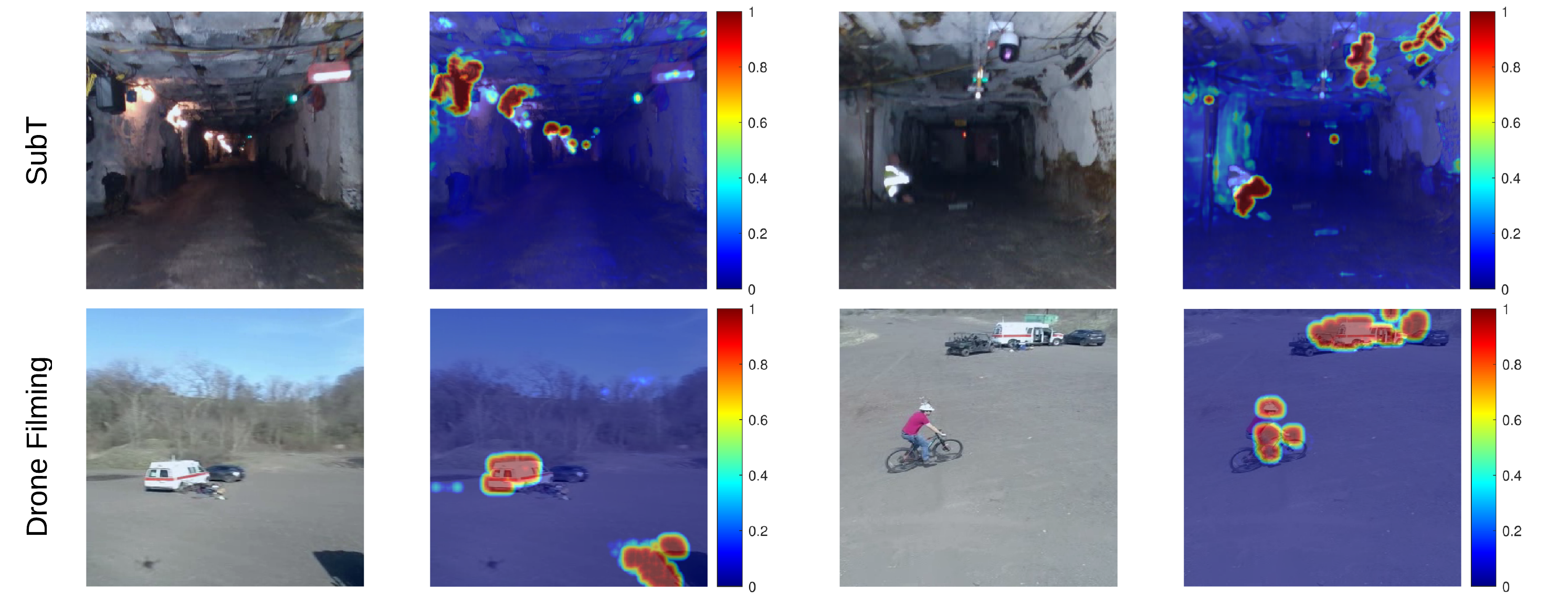}
    \caption{Selected density maps for the SubT and Drone Filming datasets. Interesting objects are correctly identified when their first appear.}
    \label{fig:heatmap}
\end{figure*}

\subsubsection{Effect of Writing Sparsity}\label{sec:effect-writing}
To show its effectiveness, we replaced our writing protocol with the one used in \cite{graves2014neural,graves2016hybrid}, which doesn't introduce the tangent operator and is denoted as `without (w/o) sparsity' in the first row of \tref{tab:overall}.
Our model achieves an average of 21.7\%, 20.7\%, and 11.1\% higher performance than the model without our sparsity operator in the normal category for $\delta=$ 1, 2, and 4, respectively.
In the difficult category, our model still has a higher overall performance of 15.5\%, 21.5\%, and 24.0\% for $\delta=$ 1, 2, and 4, respectively, which further verifies the effectiveness of the sparsity operator.

\subsubsection{Effect of Translational Invariance}\label{sec:effect-translation}
Without the invariance to large translations, the performance could drop a lot, as translational movement is very common in robotic applications.
As can be seen in the second row of \tref{tab:overall}, in which our algorithm achieves an average of 32.4\%, 33.0\%, and 20.5\% higher performance than the one without translational invariance (w/o invariance) in the normal category for $\delta=$ 1, 2, and 4, respectively.
In the difficult category, it achieves a higher margin of 20.3\%, 32.4\%, and 38.4\% for $\delta=$ 1, 2, and 4, respectively.
We notice that the performance gain has the largest margin compared to other cases, which further verifies the necessity of the introduced translational invariance.

\subsubsection{Effect of Short-term Learning}\label{sec:effect-shortterm}
Short-term learning plays an important role in quick robot deployment. The performance can be largely improved if some uninteresting scenes are known before a mission. It can be seen in the fourth row of \tref{tab:overall} that our model achieves an average of 14.6\%, 12.9\%, and 4.4\% higher accuracy than the one without short-term learning (w/o short-term) in the normal category for $\delta=$ 1, 2, and 4, respectively.
Similarly, it achieves a higher margin of 8.6\%, 13.8\%, and 14.2\% in the difficult category, respectively.
The short-term learning improves the performance only at the cost of hundreds of negative-labeled samples and several minutes of training time, which verifies the flexibility of this pipeline.

\begin{figure}[!t]
    \centering
    \includegraphics[width=1\linewidth]{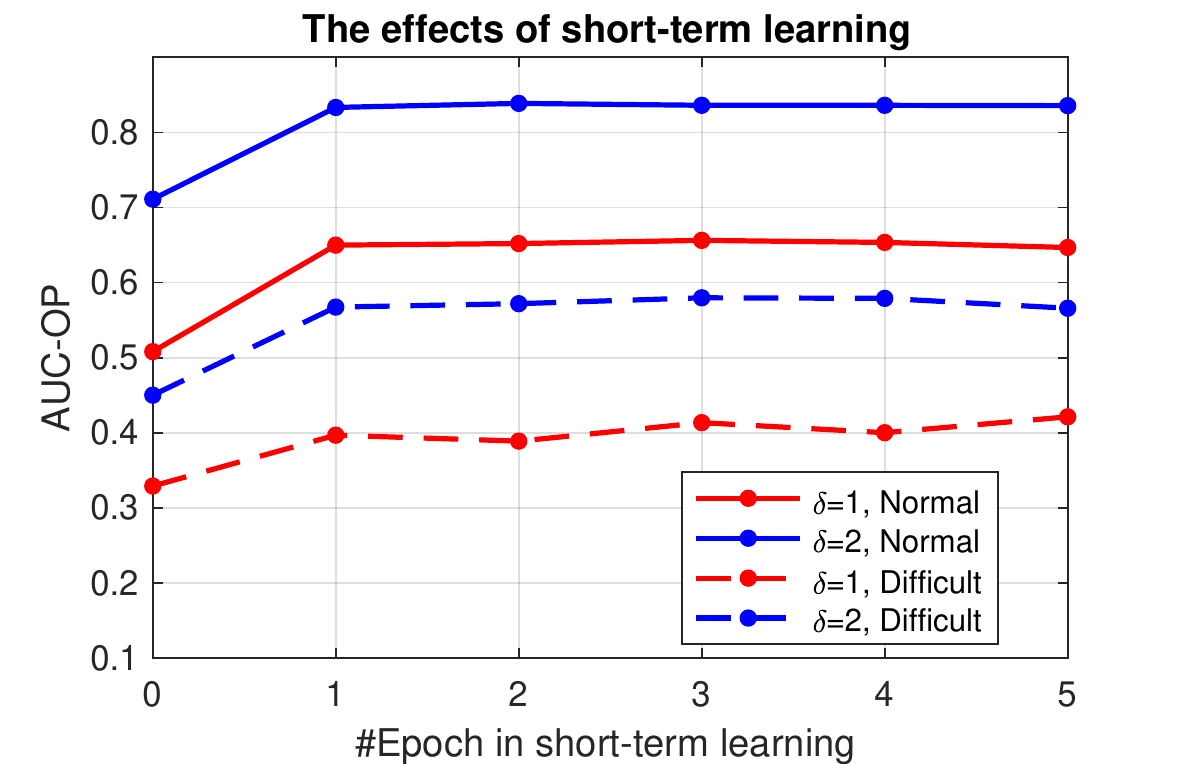}
    \caption{\fix{The final online performance is insensitive to the number of epochs in short-term learning, which indicates its efficiency and flexibility.}}
    \label{fig:short-term-epoch}
\end{figure}

\fix{
    As mentioned in \sref{sec:short-term}, we early stop the short-term learning based on either of the visual quality and the loss of reconstruction to speed up the robot deployment, hence the sensitivity of final performance (online learning) to the number of training epochs in short-term learning is also an important indicator of performance.
    To illustrate this effect, we show the performance of \ref{eq:auc-op} for models with different short-term learning time in \fref{fig:short-term-epoch}.
    It can be seen that the model performance is stable with respect to the epoch number, which indicates that the memory learning is fast, and the short-term learning generalizes well to the final performance. In the experiments, we find that the model generally produces the best trade-off performance for 3 epochs, while the performance won't be reduced dramatically if we early stop it after 1 epoch, which is able to speed up the deployment process.
}

\subsection{Density Map}\label{sec:density-map}
Recall that in online learning, the memory reading confidence is low when new objects or scenes appear suddenly because they cannot be recollected by the memory. In this sense, we are able to draw a difference map (density map) for the recalled scenes, so the interesting objects may be able to be masked by pixels with high differences.
Some examples from the SubT and Drone Filming datasets are presented in \fref{fig:heatmap}, where the map colors are normalized for better visualization.
It can be seen that some interesting objects (seen by the first time) are successfully localized such as the surveillance camera, survivor, ambulance, cyclist, \etc
Note that in this paper, we don't focus on the localization of those objects, but only focus on the interestingness of an entire frame.
We cannot guarantee that the interesting objects will always be masked by the differences of memory reading, as we expect that the interest in repetitive objects to be lost online.
In practice, we take the density maps as a secondary indicator during mission execution.

\begin{table}[t]
    \centering
    \begin{threeparttable}
        \fix{
        \caption{Performance on the underwater exploration.}
        \label{tab:scott-reef}
        \begin{tabular}{C{0.15\linewidth}C{0.13\linewidth}C{0.13\linewidth}C{0.13\linewidth}C{0.13\linewidth}}
            \toprule
            \multirow{2}{*}{Metric} & \multicolumn{2}{c}{AUC-PR$^\dagger$} & \multicolumn{2}{c}{Precision} \\
            & MemAE & Ours  & MemAE  & Ours \\
            \midrule
            Performance & 0.7595 & \textbf{0.9070}  & 0.7694 & \textbf{0.9066} \\
            \bottomrule
        \end{tabular}
        \begin{tablenotes}[normal,flushleft]
            \item  $^\dagger$The Area Under the Curve of Precision-Recall.
        \end{tablenotes}
        }
    \end{threeparttable}
\end{table}

\begin{figure}[!t]
    \centering
    \includegraphics[width=1\linewidth]{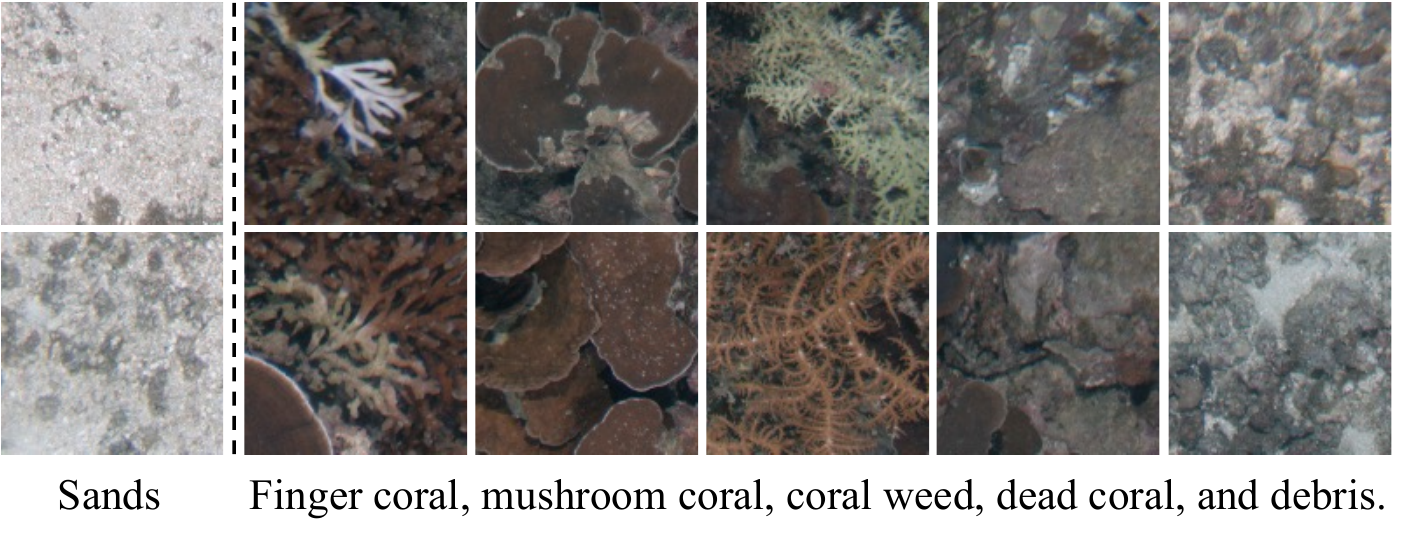}
    \caption{\fix{To test the generalization ability, we train our memory on sand images in short-term learning and then detect coral reefs in real-time.}}
    \label{fig:reef}
\end{figure}

\fix{
\subsection{Generalization Ability of Visual Memory}

In this section, we test the generalization ability of the proposed visual memory by applying it to an underwater robot exploration task.
The robots in this task are expected to collect scientifically relevant data such as coral reef images for environmental monitoring.
To this end, we conduct the short-term learning on sand images from the Scott Reef 25 dataset \cite{scottreef}, which is recorded by the Sirius autonomous underwater vehicle (AUV).
The trajectory densely covers an area of $75\metre \times 50\metre$ and consists of 50 parallel tracks and one perpendicular path across the center of the path.

Since this dataset only requires classification and doesn't require online adaptation, we remove the memory writing during online learning and report the averaged precision at a recall rate of 50\% and the area under the curve of precision-recall (AUC-PR) in \tref{tab:scott-reef} together with another memory-based method, MemAE \cite{gong2019memorizing}.
Our method produces a higher performance of 14.75\% and 13.72\% than MemAE in terms of AUC-PR and precision, respectively.
We also show several examples of the sand and coral reefs detected by our model in \fref{fig:reef}.
It can be seen that our memory-based method can correctly select coral reef images against the sand images, which verifies its generalization ability.
}

\begin{figure}[!t]
    \centering
    \includegraphics[width=1.0\linewidth]{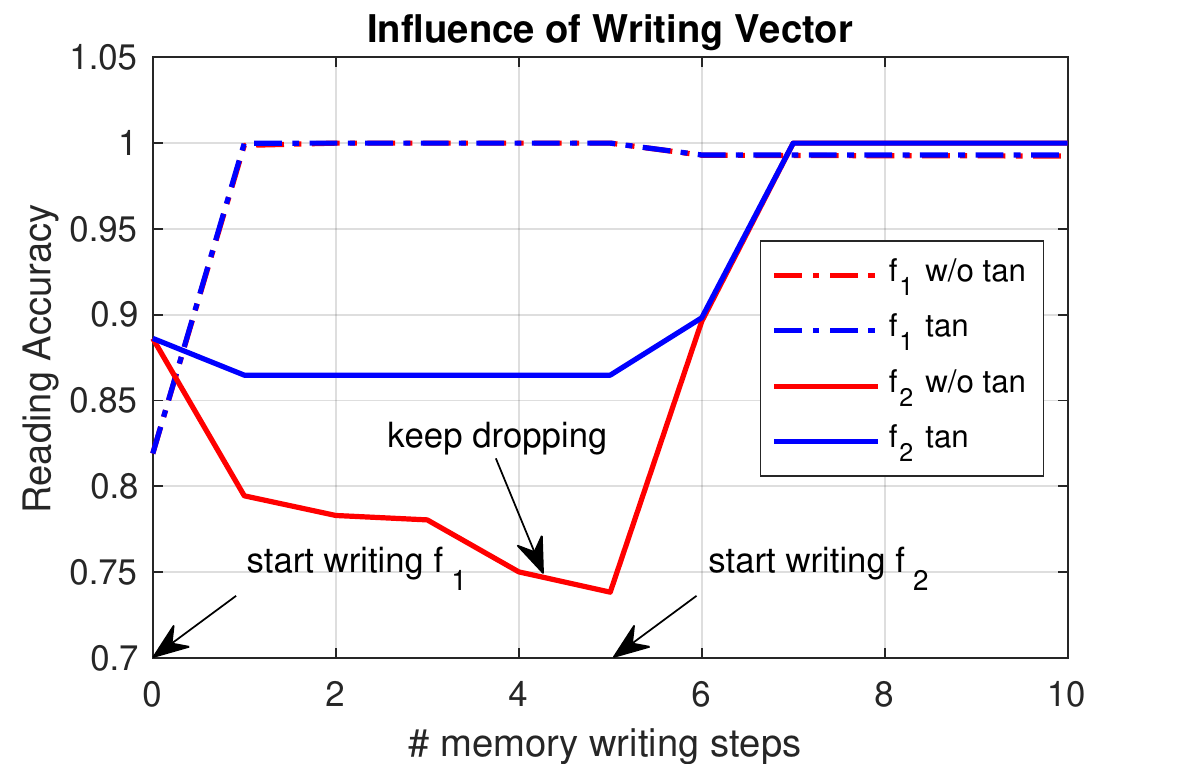}
    \caption{Influence of writing vector. Less memory cubes are affected in learning due to the  sparsity introduced by a tangent operator.}
    \label{fig:writing-vector}
\end{figure}

\section{Ablation Study}\label{sec:ablation}

In this section, we further test our algorithm and provide intuitive explanations for the influences of the writing protocol, memory capacity, translational invariance, and the ability to lose interest.
Following the ablation principle, all configurations in this section are kept the same unless stated otherwise.

\subsection{Writing Protocol}\label{sec:writing-vector}

It has been pointed out that memory learning is highly dependent on the writing vector in \eqref{eq:writing-vector}, where a tangent operator is introduced for writing sparsity.
This section explores its effect and compares it with the writing vector \eqref{eq:writing-vector-gamma} used in \cite{graves2014neural,graves2016hybrid}. Note that memory defined in \cite{graves2014neural} is vectors, thus it is not invariant to large translation.
Following the ablation principle, we use the same 4-D memory structure and the translation-invariant reading protocol proposed in this paper.
\begin{equation}\label{eq:writing-vector-gamma}
\mathbf{w} = \softmax(\gamma \cdot D(\mathbf{x}, \mathbf{M})),
\end{equation}
where $\gamma$ is a parameter related to temperature.
To show the writing performance, we write two random 3-D tensors into the memory, \ie $\mathbf{f}_1$ and $\mathbf{f}_2$, and compare their reading accuracy in terms of the cosine similarity defined in \eqref{eq:reading-accuracy}.
\begin{equation}\label{eq:reading-accuracy}
S^c(\mathbf{r}, \mathbf{f}) = \frac{\sum(\mathbf{r}\odot \mathbf{f})}{\|\mathbf{r}\|_\mathbf{F}\cdot\|\mathbf{f}\|_\mathbf{F}},
\end{equation}
where $\mathbf{r}$ and $\mathbf{f}$ are the memory reading and writing tensors consecutively.
In experiments, we set $\gamma_w=\gamma_r=5$ and write both $\mathbf{f}_1$ and $\mathbf{f}_2$ 5 times continuously and show their reading accuracy in terms of number of writing in \fref{fig:writing-vector}.

\begin{figure}[!t]
    \centering
    \includegraphics[width=1.0\linewidth]{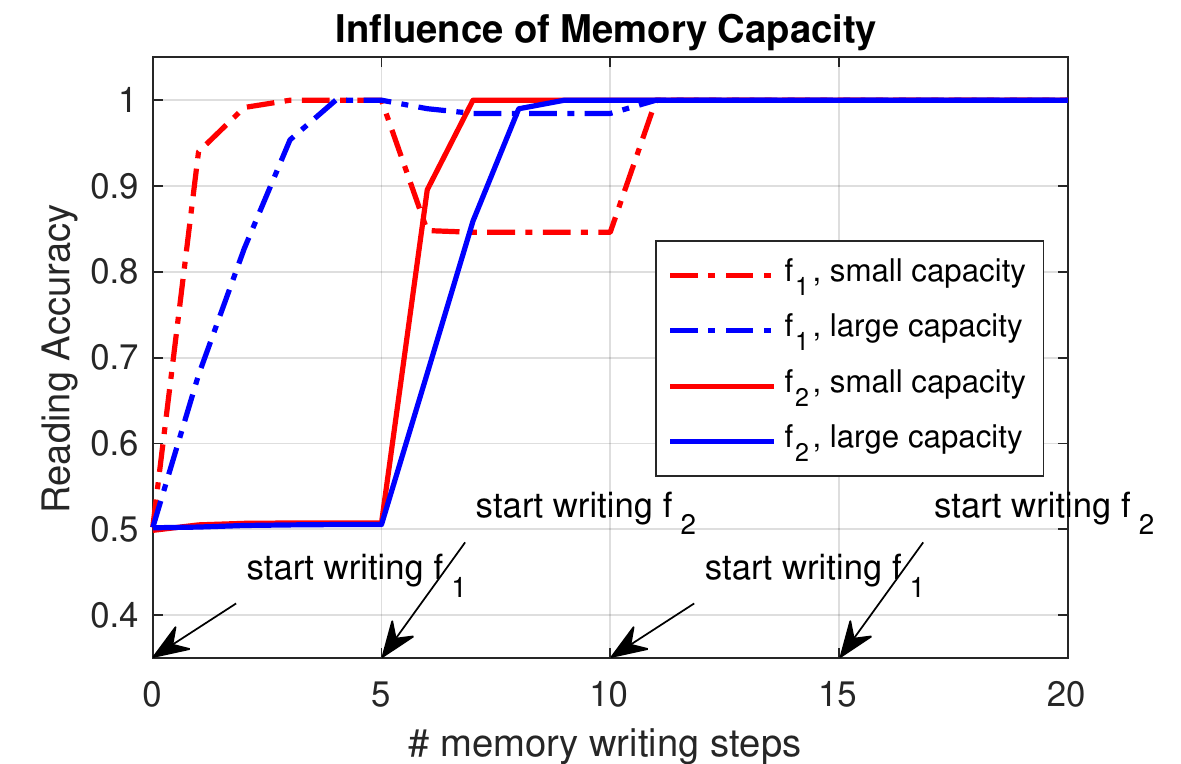}
    \caption{Influence of memory capacity. Larger memory capacity leads to more computation but easier incremental learning.}
    \label{fig:memory-capacity}
\end{figure}

It can be seen that both memories are able to remember the random tensors after repeatedly writing. However, when writing a vector without a tangent operator, the accuracy of $\mathbf{f}_2$ keeps dropping even when $\mathbf{f}_1$ is learned, \ie $S^c(\mathbf{r}_1, \mathbf{f}_1)\approx1$. This is because all memory cubes are affected due to the non-sparse writing vector in \eqref{eq:writing-vector-gamma}.
This will be a severe issue when a robot keeps learning the same thing (observing the same scene), since the learned knowledge may be forgotten due to the non-sparse writing.
Nevertheless, our writing vector with the tangent operator is able to map the weight of $\mathbf{f}_1$ to infinite when $\mathbf{f}_1$ is learned, resulting in safer memory writing as only a few memory cubes are affected. This further verifies the effectiveness of the new writing vector.

\subsection{Memory Capacity}\label{sec:capacity}

This section explores the effects of memory capacity, \ie the number of memory cubes $n$, which is an important hyper-parameter for incremental learning.
Similarly, we write two same random 3-D tensors $\mathbf{f}_1$ and $\mathbf{f}_2$ five times sequentially into two different memories in terms of the memory capacity.

As can be seen in \fref{fig:memory-capacity}, both memories are able to learn random samples in a short time, while the reading accuracy of $\mathbf{f}_1$ drops a lot for smaller capacity after starting to write $\mathbf{f}_2$, although it is remembered later when $\mathbf{f}_1$ is written again.
We observe a similar phenomenon when the number of samples is around the same or larger than the memory capacity.
This means that memory with a small capacity quickly forgets old knowledge when learning new knowledge. While for larger capacity, reading accuracy can be less affected by new knowledge, resulting in safer and easier incremental learning, which is vital for real-time robotic applications.
We can leverage this property for model design since uninteresting objects can also become interesting in some cases.

\begin{figure}[!t]
    \centering
    \includegraphics[width=1.0\linewidth]{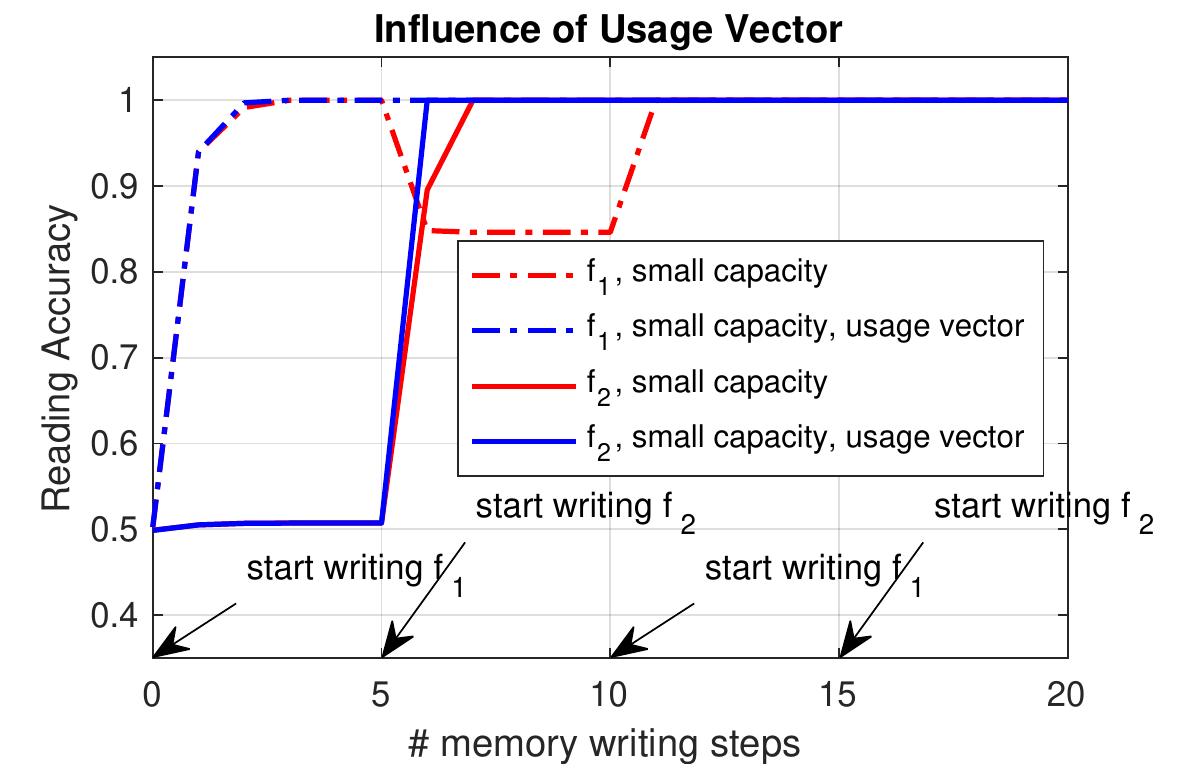}
    \caption{Influence of usage vector. Even the memory with a small capacity become easier for incremental learning due to the usage vector.}
    \label{fig:usage}
\end{figure}

\subsection{Memory Usage Vector}\label{sec:usage-vector}

As mentioned in \sref{sec:efficiency}, we achieve more than 3 times faster than the previous version \cite{wang2020visual} without sacrificing its accuracy.
This is because the memory usage vector \eqref{eq:memory-usage} improves the efficiency of space usage, allowing us to use a smaller memory capacity.
However, as mentioned in \sref{sec:capacity}, a small memory capacity means that it's easier to forget old knowledge when learning new knowledge.
This section provides an intuitive explanation for this.
To this end, we perform the same experiments described in \sref{sec:capacity} and introduce the usage vector to the setting with small capacity.

As can be seen in \fref{fig:usage}, the forgetting problem of small capacity with the usage vector is dramatically alleviated, as $\mathbf{f}_1$ is not forgotten when learning $\mathbf{f}_2$, especially compared to \fref{fig:memory-capacity}.
Therefore, we can use a smaller memory to achieve similar performance compared to the previous version, resulting in a much faster running speed.
Note that this is a substantial improvement, as we can easier apply this method to robots with ultra-low power processors, \eg UAV.

\subsection{Translational Invariance}\label{sec:translation-invariance}

\begin{figure}[!t]
    \centering
    \includegraphics[width=1.0\linewidth]{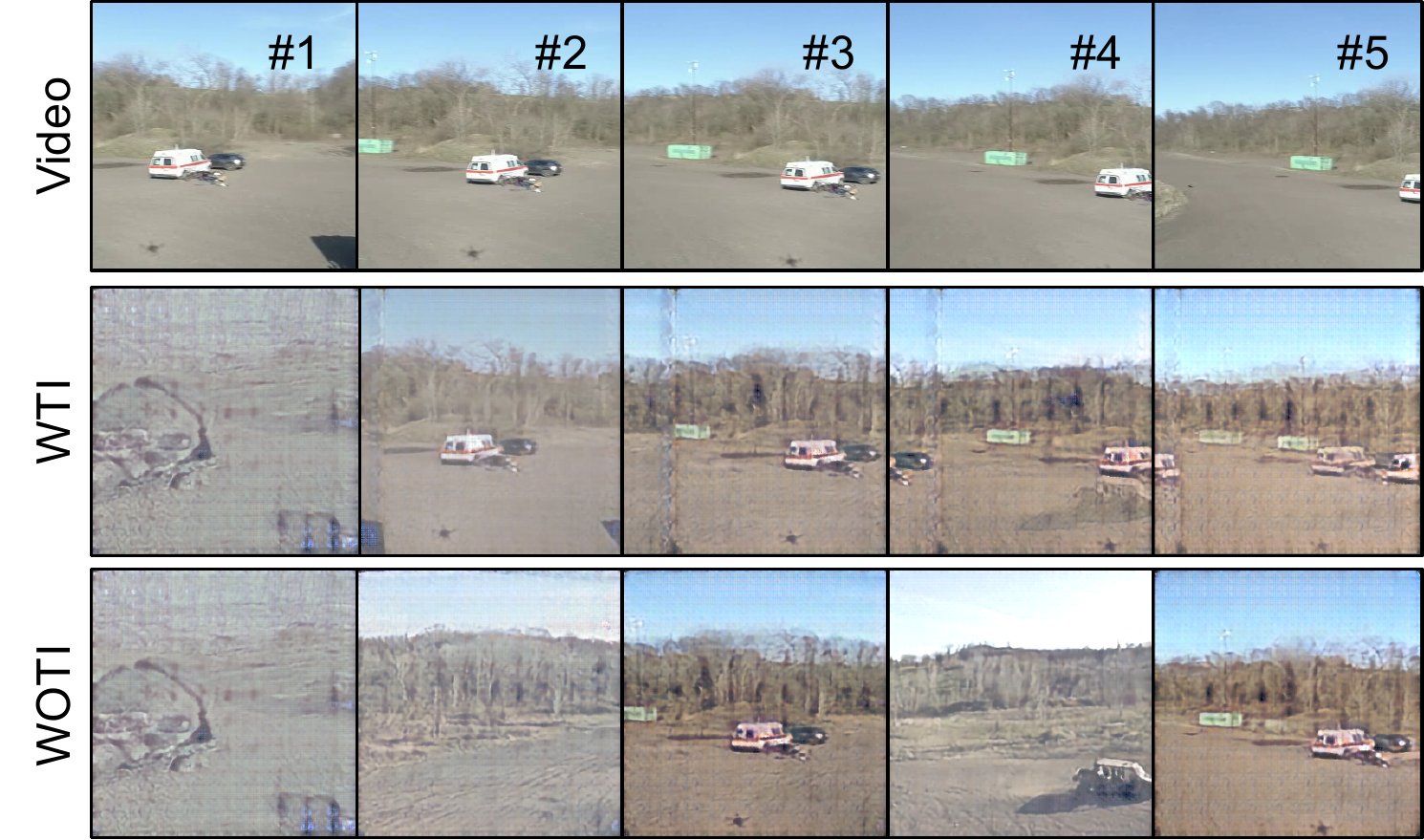}
    \caption{Memory reading with translational invariance (WTI) recall translational scenes is better than the one without invariance (WOTI). The sequence is trimmed from the Drone Filming dataset \cite{wang2019improved} and contains five frames where an ambulance appears and disappears in the first and last frame, sequentially. The cross-correlation similarity and CNN features contribute complete translational invariance to memory recall.}
    \label{fig:translational-invariance}
\end{figure}

It has been mentioned that human beings have a great capability of recalling memory when a similar scene is observed before.
Although CNN features are invariant to small translations \cite{wang2019kervolutional}, they still fail to recall memory when large translations occur.
To solve this problem, we introduce translation-invariant reading vector by cross-correlation in \eqref{eq:cross-correlation}.
We next test it on the Drone Filming dataset \cite{wang2019improved} in \fref{fig:translational-invariance}. In this sequence, an ambulance appears suddenly in the 1st frame and disappears in the 5th frame.
We construct two memory modules to learn this video based on the online learning strategy presented in \sref{sec:online}.
The first module adopts the cross-correlation similarity presented in \sref{sec:memory-reading} for memory reading (denote as WTI), while another one adopts the cosine similarity (WOTI).
It can be seen that both modules cannot recall the memory for the 1st frame, since the ambulance is not seen before.
However, the module WTI is able to recall the memory precisely in the subsequent frames, while the module WOTI quickly fails, although its reading is still meaningful, \eg the 2nd and 4th frames have correct patterns for sky, trees, and ground.
The recalled memory for the 3rd frame from module WTI is roughly a translated replica of the 2nd frame of the video (this also occurs at the 4th and 5th frame), which means that WTI correctly takes the 2nd frame as the most similar scene to the 3rd frame.
This phenomenon verifies the translational invariance of our reading protocol.

Note that there is a small translation between the video and the 2nd frame from WTI.
This is because the invariance to small translations of CNN features, \ie  the features look the same for visual memory, although they appear with a small translational difference.
Therefore, the introduced cross-correlation similarity together with the CNN features contribute complete invariance of translation to memory recall.

\begin{figure*}[ht]
    \centering
    \includegraphics[width=1.0\linewidth]{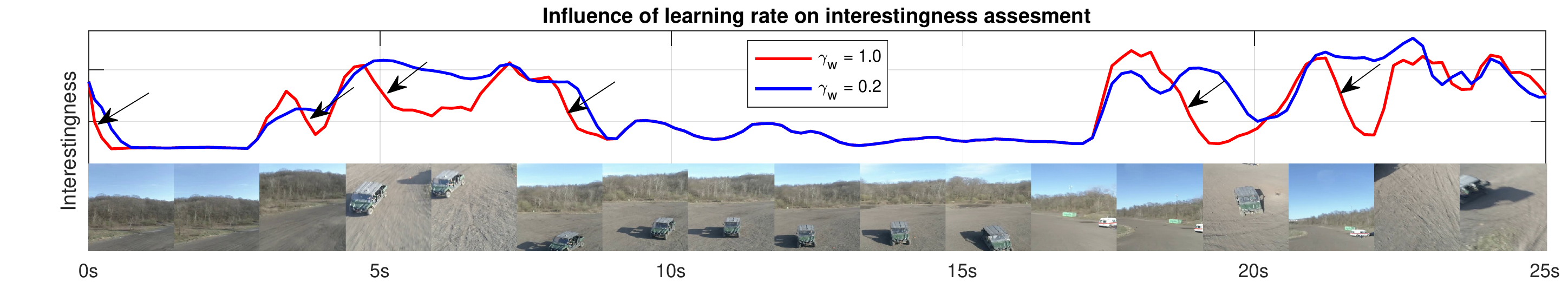}
    \caption{Visual interestingness with different writing rates for drone video footage \cite{wang2019improved}. As indicated by the arrows, a larger writing rate results in a faster loss of interest for new objects or scenes during online learning.}
    \label{fig:drone-filming}
\end{figure*}

\subsection{Loss of interest}\label{sec:losing-interest}

To test the online learning capability of losing interest for the algorithm, we perform a qualitative test based on the Drone Filming dataset.
The objects tracked in this dataset, \eg cars or bikes, are relatively stable with respect to the drone, while the background of the tracked objects keeps changing due to their movements.
This makes it suitable for testing the capability of online learning.
One of the video clips is shown in \fref{fig:drone-filming}, where two different online learning speeds are adopted, \ie $\gamma_w = 1.0$ and $\gamma_w =0.2$.
It can be seen that the interestingness levels of both settings become high when new objects or scenes appear, which means that both settings are able to detect novel objects.
However, the interestingness level with a larger writing rate always drops faster, meaning that it is quicker to lose interest in similar scenes.
This verifies our objective that a simple hyper-parameter can be adjusted for different missions.
We notice that in \fref{fig:memory-capacity}, the memories with different capacities have different learning speeds. It can also be adjusted by the writing rates tested in this section.

\fix{
\section{Limitation \& Discussion}

We mainly researched the topic of general interestingness recognition for field robots that often operate in unknown and unstructured environments where interesting objects and scenes are critical for robotic tasks such as search and rescue. However, for other cases such as household robots that often operate in known environments, the potential interesting objects could have different and even multiple definitions in tasks such as cleaning and organizing, where our algorithm may not be applicable.
For example, we may be interested in a broom during cleaning, while interested in the entire messy room when reorganizing it.
In the future, we will consider the scenario that only specific objects are interesting but not predefined in a robotic mission.
To achieve this, we will take users' commands into account for better adapting to the new environments.
The model needs to be updated in real-time and search for similar interesting objects/scenes via online learning from a few labeled examples selected by the user during mission execution.
This could further benefit the robotic tasks such as manipulation, exploration, and decision-making.
}

\section{Conclusion}
In this paper, we developed an unsupervised online learning algorithm for robotic visual interestingness recognition.
We first introduced a novel translation-invariant 4-D visual memory, which can be trained without the back-propagation algorithm.
We also introduced a usage vector into the memory to improve the efficiency of space usage.
To better fit for practical applications, we designed a three-stage learning architecture: long-term, short-term, and online learning.
The long-term learning stage is responsible for human-like, real-life knowledge accumulation and trained on unlabeled data via back-propagation.
Short-term learning is responsible for learning environmental knowledge and trained via visual memory for quick robot deployment, while online learning is responsible for environment adaption and leverage visual memory to identify interesting scenes.
The experiments show that being implemented on a single machine, our approach is able to learn online and find interesting scenes efficiently in real-world robotic tasks.
It is also shown that our approach even outperforms supervised methods with a large margin on the SubT dataset.
We expect that it will play an important role in robotic applications such as exploration and decision-making.

\section*{Acknowledgments}
This work was partially sponsored by the ONR grant \seqsplit{\#N0014-19-1-2266} and ARL DCIST CRA award \seqsplit{W911NF-17-2-0181}. The human subject survey was approved under \seqsplit{\#2019\_00000522}. 
The authors would like to thank all members of the Team Explorer for their constructive suggestions and for providing data collected from the DARPA Subterranean Challenge Tunnel Circuit.
\fix{The authors would also like to acknowledge the Australian Centre for Field Robotics' marine robotics group for providing the data of \textit{Scott Reef 25}.}

\balance
{
    \small
    \bibliographystyle{IEEEtran}
    \bibliography{IEEEabrv,egbib,papers,mypublication}
}

\ifCLASSOPTIONcaptionsoff
\newpage
\fi

\end{document}